\documentclass{article}

\usepackage{microtype}
\usepackage{graphicx}
\usepackage{subcaption}
\usepackage[colorinlistoftodos, textsize=tiny]{todonotes}
\usepackage{booktabs} 
\usepackage{multicol}

\usepackage{hyperref}
\usepackage{fontawesome5}

\usepackage[accepted]{icml2026}

\usepackage{amsmath}
\usepackage{amssymb}
\usepackage{mathtools}
\usepackage{amsthm}
\usepackage{multirow}
\usepackage{algorithm}
\usepackage{algorithmic}

\usepackage[capitalize,noabbrev]{cleveref}

\theoremstyle{plain}

\theoremstyle{definition}

\theoremstyle{remark}

\icmltitlerunning{Sparse Autoencoders for Interpretable Emotion Control in Text-to-Speech}

\begin{document}

\twocolumn[
  \icmltitle{Sparse Autoencoders for Interpretable Emotion Control in Text-to-Speech}



  \icmlsetsymbol{equal}{*}

\begin{icmlauthorlist}
    \icmlauthor{Hongfei Du}{wm}
    \icmlauthor{Jiacheng Shi}{wm}
    \icmlauthor{Sidi Lu}{wm}
    \icmlauthor{Gang Zhou}{wm}
    \icmlauthor{Ye Gao}{wm}
\end{icmlauthorlist}

  \icmlaffiliation{wm}{Department of Computer Science, William \& Mary, USA}

  \icmlcorrespondingauthor{Ye Gao}{ygao18@wm.edu}
  \icmlcorrespondingauthor{Hongfei Du}{hdu02@wm.edu}

  \vskip 0.3in
]



\printAffiliationsAndNotice{}  

\begin{abstract}
Integrating large language models (LLMs) into text-to-speech (TTS) systems has improved speech expressiveness, yet interpretable emotional control remains challenging.
Existing approaches primarily rely on external conditioning or global activation steering, offering limited insight into the internal representations underlying emotional control.
In this work, we analyze emotion-related variation in the semantic hidden states of LLM-based TTS models using sparse autoencoders (SAEs) to identify sparse latent features. Our analysis shows that emotional variation is distributed across multiple sparse latent features, while intervening on a small subset enables interpretable emotion control.
Building on this observation, we introduce a feature-level intervention framework for bidirectional emotion induction and suppression without modifying backbone parameters.
We further show that distinct latent features are associated with specific acoustic attributes (e.g., pitch), suggesting that emotional expression arises from coordinated latent contributions rather than a single global shift. Empirically, steering these sparse latent features achieves comparable or superior emotion induction and suppression performance relative to global steering and existing TTS baselines. \href{https://maxpeaced.github.io/sae-demo/}{\faGithub\ GitHub-Demo}


\end{abstract}

\begin{figure}[ht]
  \centering
  \includegraphics[
  width=1\columnwidth,
  trim=0cm 0cm 0cm 0cm,
  clip
]{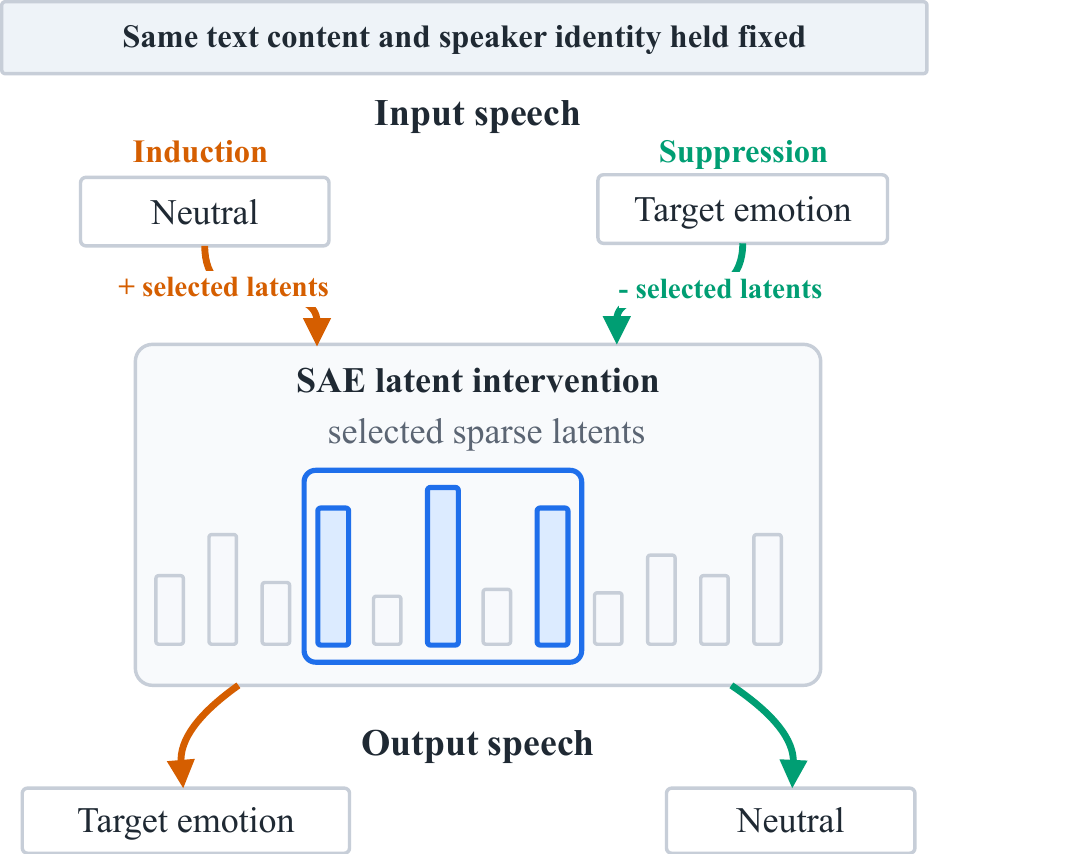}
 \caption{
  Conceptual overview of SAE-based bidirectional emotion control in the semantic backbone. Holding text content and speaker identity fixed, we intervene on selected SAE latent features. Increasing these features induces a target emotion from neutral speech, whereas decreasing them suppresses the target emotion toward neutral speech.}
  \label{fig:intro-concept}
\end{figure}

\section{Introduction}
LLM-based text-to-speech (TTS) models, which integrate large language models into the speech generation pipeline, have recently demonstrated strong capability in producing natural and expressive speech \cite{du2025cosyvoice3, zhou2025indextts2, wang2025spark, zhang2023speak}. Such expressiveness is critical for real-world applications, including audiobooks \cite{guo2024text} and human-computer interaction \cite{wadley2022future}. As a result, controllable emotional expression has become increasingly important in modern TTS systems.

Existing emotion-controllable TTS methods typically rely on either label-based or reference-based conditioning. Label-based approaches use discrete emotion categories or textual prompts \cite{kim2021expressive, yang2025emovoice, ji2024textrolspeech, liu2025uddetts}. In these methods, emotions are modeled as predefined classes that tend to average out nuanced variations and limit expressive diversity. Reference-based methods transfer style from reference audio \cite{wang2025spark, ji2025controlspeech}, enabling natural prosody transfer while requiring suitable reference samples and providing only instance-level, non-adjustable control. Neither paradigm is inherently interpretable, and both offer limited flexibility once the input conditions are fixed.
Recent work explores activation steering for training-free emotion control in TTS \cite{xie2025emosteer}, directly modifying hidden states during inference within a diffusion transformer. 
However, since emotional expression in speech emerges from multiple acoustic factors (e.g., pitch, energy, and prosody), steering with a single dense mean-difference direction captures only a global representation shift rather than separable feature-level variation, thereby limiting interpretability and modular control. This motivates us to investigate whether emotion-related variation can be decomposed into sparse latent features within the semantic backbone.

In this work, we analyze emotional variation in the semantic hidden representations of LLM-based TTS models, focusing on how emotion-related signals are organized across internal activations rather than as global hidden-state shifts.
To analyze this internal representation space, we train a sparse autoencoder (SAE) \cite{olshausen1997sparse} on hidden activations collected during speech semantic-token generation in an unsupervised manner to learn sparse latent features prior to acoustic synthesis.

Our analysis in Section~\ref{interpret} identifies emotion-related latent features within the semantic backbone. We further show that individual latent features are associated with distinct acoustic variations, such as changes in pitch and spectral brightness. As illustrated in Figure~\ref{fig:intro-concept}, intervening on these latent features enables both emotion induction and suppression without modifying backbone parameters. In summary, our main contributions are as follows:
\begin{itemize}
\item We demonstrate that sparse autoencoders (SAEs) extract interpretable emotion-related features from the semantic backbone of LLM-based TTS models during semantic-token generation.

\item We show that emotional expression in TTS is encoded across multiple sparse latent features rather than along a single global direction, offering a representation-level perspective on emotion modeling in speech synthesis.

\item Building on these findings, we propose a feature-level emotion control framework for LLM-based TTS that enables bidirectional emotion induction and suppression without modifying backbone parameters.
\end{itemize}

\section{Related Work}
\paragraph{Sparse Autoencoders (SAEs)}
Sparse Autoencoders (SAEs) decompose polysemantic neural representations into sparse, interpretable features \cite{bricken2023monosemanticity, yun2021transformer, elhage2022toy}.
They have been increasingly applied in large language models (LLMs) \cite{cunningham2023sparse,ferrando2024know} and vision--language models (VLMs) \cite{lou2025sae} for representation steering \cite{templeton2024scaling, durmusevaluating, joseph2025steering} and concept unlearning \cite{cywinski2025saeuron}.
Prior work has applied SAEs to audio generative models, primarily analyzing latent features of acoustic autoencoders and mapping them to acoustic properties such as pitch, amplitude, and timbre \cite{paek2025learning}. This approach focuses on interpreting acoustic latent spaces, whereas we apply SAEs to the semantic backbone of autoregressive LLM-based TTS models to investigate emotion-related latent features during speech semantic-token generation. In doing so, we shift interpretability from acoustic realization to high-level emotional structure before acoustic synthesis.

\paragraph{Emotion-Controllable Text-to-Speech}
Controllable emotional TTS aims to generate speech with specified affective characteristics \cite{kim2021expressive}. Existing approaches are commonly divided into label-based and reference-based methods.
Label-based methods regulate emotion through predefined discrete categories or textual descriptions. Early work conditions Tacotron-based models on discrete emotion labels \cite{lee2017emotional}, while more recent LLM-based systems such as EmoVoice \cite{yang2025emovoice} support flexible control via natural language prompts. Dimensional extensions further introduce continuous affective coordinates (e.g., ADV space) to enhance controllability \cite{liu2025uddetts}. These approaches primarily operate through external conditioning signals defined in advance.
Reference-based methods instead transfer speaking style from reference audio \cite{wang2018style,skerry2018towards,jia2018transfer}, enabling flexible expression without explicit labels. However, the learned style representations are typically opaque and require selecting suitable reference speech at inference time.
More recently, activation steering has been explored to enable emotion control by directly modifying hidden states during inference \cite{xie2025emosteer}. This approach performs steering in the original hidden space and typically uses a dense mean-difference direction to induce global representation shifts. In contrast, we investigate whether emotion-related variation can be identified and modulated via sparse latent features within the semantic backbone.

\begin{figure*}[ht]
  \centering
  \includegraphics[
      width=1.\textwidth,
      trim=0cm 1.3cm 0.3cm 0cm,
      clip
  ]{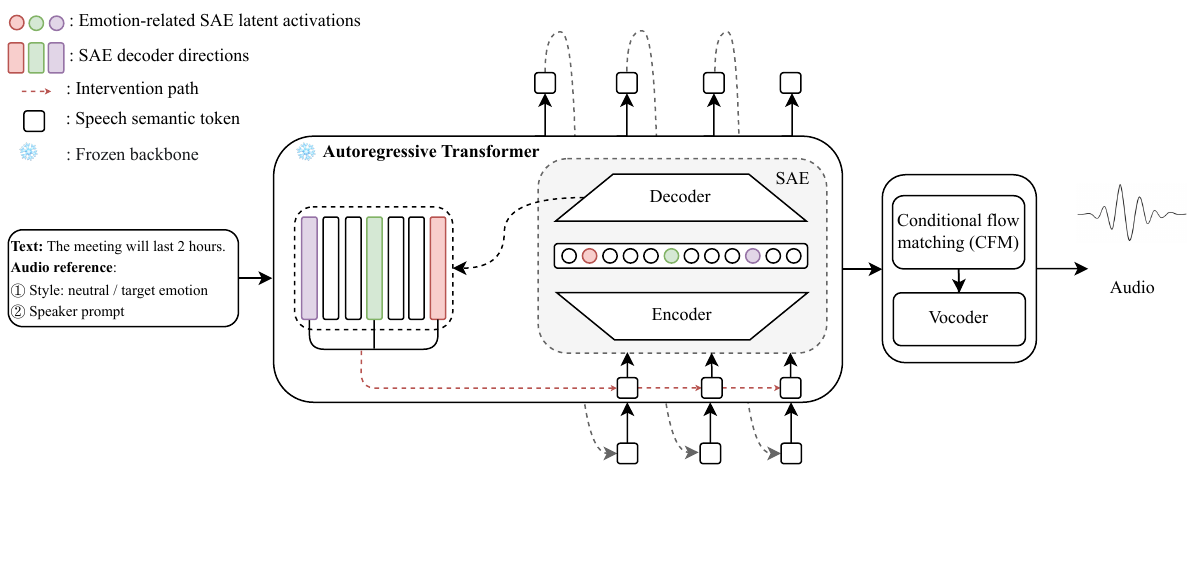}
 \caption{
 Overview of SAE-based emotion modulation in an LLM-based TTS model. 
The semantic backbone generates token-level hidden representations conditioned on text and optional audio references. 
A sparse autoencoder (SAE) maps residual-stream activations to sparse latent activations.
Selected emotion-related latent features are modulated through the intervention path and decoded back into the residual stream.
The resulting representations are passed to the CFM module and vocoder for acoustic synthesis.
} 
  \label{fig:sae-tts}
\end{figure*}

\paragraph{Activation Steering and Linear Representations}
Activation steering modulates internal representations during inference to control model behavior without additional training. It has been applied to influence behaviors such as truthfulness \cite{li2023inference,wang2025adaptive}, safety \cite{liu2023context}, and stylistic attributes \cite{subramani2022extracting} in large language models, and has been extended to vision--language models \cite{gavrikov2024can,joseph2025steering} and text-to-image generative models \cite{kim2025concept,li2024get}. 
Recent hybrid TTS architectures integrate autoregressive semantic modeling with flow-based/diffusion-based acoustic generators \cite{zhou2025indextts2,deng2025indextts,du2024cosyvoice1,du2024cosyvoice2}, enabling instruction-level control via natural language prompts. In TTS, prior steering work has explored interventions in the diffusion-transformer acoustic generation stage \cite{xie2025emosteer}. However, emotional variation may originate upstream in the autoregressive semantic backbone. If emotion is encoded in these semantic representations, intervening at this stage provides a complementary approach to emotion control. We therefore use sparse autoencoders to identify and modulate interpretable emotion-related latent features within the autoregressive semantic backbone.

\section{Methodology}

\subsection{Sparse Autoencoders for LLM-based TTS Models}
\label{intro-sae}
Rather than modifying downstream acoustic modules (e.g., diffusion- or flow-based synthesis), we integrate a sparse autoencoder (SAE) into the semantic backbone of the autoregressive LLM-based TTS model, as illustrated in Figure~\ref{fig:sae-tts}. By intervening prior to acoustic synthesis, our approach directly modulates SAE latent features derived from the semantic representations passed to the acoustic generator.

During semantic-token generation, the autoregressive transformer produces hidden representations conditioned on text and optional reference inputs. We intercept the layer-$l$ residual stream and extract activations $x \in \mathbb{R}^d$ at each generated semantic-token position, where $d$ denotes the hidden dimensionality of the semantic backbone. These residual activations are typically dense and entangled, making direct manipulation less suitable for feature-level emotion control.

The SAE represents each dense residual activation $x$ using an overcomplete set of learned latent features. At each token position, the encoder produces a sparse latent activation vector $z$ with only a small number of active latent features. This~sparse feature representation provides an interface for identifying interpretable emotion-related and prosodic patterns and intervening on the corresponding latent features. During inference, we modulate selected latent features, decode the modified activations back into residual representations, and pass the resulting representations to the acoustic generator for speech synthesis.

To learn sparse latent features, we employ a $k$-sparse autoencoder \cite{makhzani2013k} that maps the input $x$ into an overcomplete latent space of dimension $n$ ($n > d$). This mapping transforms residual stream activations into sparse latent activations under a different basis. The encoder first centers the input using a learnable bias $b_{\text{pre}} \in \mathbb{R}^d$, and then applies a linear projection $W_{\text{enc}} \in \mathbb{R}^{n \times d}$.

To obtain sparse and more interpretable representations, we constrain each input to activate only a small number of latent features. Concretely, we apply a Top-$k$ operator \cite{gao2024scaling} that retains only the $k$ largest activations while suppressing the rest, introducing a competitive bottleneck among latent features:
\begin{equation}
z = \operatorname{Top}_{k} \big( \operatorname{ReLU} ( W_{\text{enc}}(x - b_{\text{pre}}) + b_{\text{enc}} ) \big),
\end{equation}
where $b_{\text{enc}} \in \mathbb{R}^n$ is the encoder bias. This operation encourages latent features to compete in explaining the input, promoting active latent features that capture more localized semantic~factors.

The decoder maps each latent feature back to a direction in the residual stream. The input is subsequently reconstructed through a linear decoder:
\begin{equation}
\hat{x} = W_{\text{dec}} z + b_{\text{pre}},
\end{equation}
where the columns of $W_{\text{dec}} \in \mathbb{R}^{d \times n}$ can be interpreted as directions in the residual space, each associated with a specific latent feature. The primary training objective is to minimize the reconstruction error $\mathcal{L}_{\text{rec}} = \|x - \hat{x}\|_2^2$, so that the latent activation vector $z$ preserves the semantic information required for downstream TTS tasks.

A critical challenge in training overcomplete SAEs is the \textit{dead latent} problem, where a subset of latent features remains inactive across the data distribution, leading to under-utilized capacity. To promote better feature utilization, we use selected inactive features to model the residual reconstruction error $(x - \hat{x})$, allowing them to capture complementary information not represented by the Top-$k$ active features. Let $\tilde{z}$ denote the activations of selected inactive latent features and $\tilde{x} = W_{\text{dec}} \tilde{z}$ denote their corresponding reconstruction. The auxiliary loss is defined as:
\begin{equation}
\mathcal{L}_{\text{aux}} = \lVert (x - \hat{x}) - \tilde{x} \rVert_2^2.
\end{equation}

We train the SAE with a reconstruction objective augmented by an auxiliary loss, $\mathcal{L} = \mathcal{L}_{\text{rec}} + \lambda_{\text{aux}}\mathcal{L}_{\text{aux}}$, which improves latent utilization and reduces dead latents.

\subsection{Sparse Latent Feature Identification and Intervention for Emotion Control}
After training the SAE on hidden representations from the LLM-based TTS semantic backbone, emotion steering proceeds in two stages. First, we identify emotion-related latent features through controlled activation analysis. Second, during inference, we intercept the layer-$l$ residual stream, encode it into sparse latent activations, modify the activations of the selected latent features, and decode the result back into the residual space. The modified representations are then passed to the CFM module and vocoder. 

\subsubsection{Identifying Emotion-Related Latent Features}
\label{main:select}
To identify emotion-related SAE latent features, we analyze activation patterns under strictly controlled emotional conditions. We fix the input text (see Appendix~\ref{app:text-prompt}) and speaker identity, varying only the emotional reference signal. In particular, neutral speech is treated as the baseline condition, and each target emotion (e.g., happiness, sadness, anger) is contrasted against the corresponding neutral condition. This controlled design reduces confounding effects from lexical content and speaker-dependent variation, making differences in latent activations attributable to emotion.

For each input sample $u$ under emotion condition $e$, the autoregressive backbone produces a sequence of residual stream activations at layer $l$, denoted as $X_l^{(e)}(u) \in \mathbb{R}^{T_u \times d}$, where $T_u$ is the number of generated semantic tokens for sample $u$. These activations are encoded by the trained SAE to obtain sparse latent activations $A^{(e)}(u) \in \mathbb{R}^{T_u \times n}$, and we denote the activation of latent feature $i$ at token position $t$ by $a_{i,t}^{(e)}(u)$.

A key challenge is to identify latent features that are consistently associated with a given emotion, rather than those that exhibit occasional spikes. Peak activations can be dominated by isolated token-level effects and may not reflect stable emotional patterns. In contrast, activation frequency reflects how reliably a latent feature participates across different contexts, making it a more stable indicator of whether a latent feature is involved in expressing a given emotion.

To aggregate activations across samples, we convert token-level activations into a sentence-level occurrence indicator: latent feature $i$ is counted as active for sample $u$ if it fires at any generated token position.
\begin{equation}
\mathbf{1}_i^{(e)}(u) 
= \mathbb{1}\left[
\exists\, t \in \{1,\dots,T_u\} 
\text{ s.t. } a_{i,t}^{(e)}(u) > 0
\right].
\end{equation}

This definition operates at the sentence level and does not differentiate between isolated and sustained token-level activations. We adopt a sentence-level metric because emotional expression is typically sustained across an utterance rather than localized to individual tokens. Moreover, each semantic token corresponds to a local generation step and only a short portion of the resulting speech, making token-level variations more susceptible to transient noise than stable emotional cues. Since emotion and neutral conditions are evaluated under identical sparsity constraints, occasional token-level spikes are less likely to bias the comparison.

We then compute the sentence-level activation rate:
\begin{equation}
    r_i^{(e)} = 
\frac{1}{|\mathcal{D}|}
\sum_{u \in \mathcal{D}}
\mathbf{1}_i^{(e)}(u).
\end{equation}

Finally, to isolate emotion-specific modulation, we compare each emotion $e$ with the neutral condition under matched text and speaker conditions. Since each target-emotion sample is paired with a corresponding neutral sample, we define the emotion selectivity score as:
\begin{equation}
\label{emo-score}
\Delta_i^{(e)} =
\frac{1}{|\mathcal{D}|}
\sum_{u \in \mathcal{D}}
\left(
\mathbf{1}_i^{(e)}(u) -
\mathbf{1}_i^{(\mathrm{neutral})}(u)
\right).
\end{equation}
Latent features with the largest positive $\Delta_i^{(e)}$ are selected as emotion-related features, as they are more frequently recruited under the target emotional condition than under the matched neutral condition. Because this score is computed as a paired emotion--neutral difference under matched text and speaker conditions, it filters out globally frequent but condition-insensitive latent features. We further compare this sentence-level selectivity criterion with magnitude-based and token-level alternatives in Appendix~\ref{app:alternative_selection_criteria}, where it improves emotion alignment for all three target emotions.

\subsubsection{Sparse Latent-Feature Steering}
\label{steer}

Given the trained SAE, our key insight is that emotion control can be interpreted as steering the residual representation along a sparse set of learned directions.
Specifically, the residual representation can be approximated by the SAE decoder as:
\begin{equation}
x \approx b_{\text{pre}} + \sum_{j=1}^{n} a_j(x)\, d_j,
\end{equation}
where $a_j(x)$ denotes the activation of latent feature $j$ and $d_j$ is the corresponding decoder direction in the residual space. Modifying $a_j(x)$ therefore steers the residual representation along $d_j$~\cite{ferrando2024know}.

Let $\mathcal{F}_e$ denote the set of emotion-related latent features identified for emotion $e$. For a residual representation $x_{l,t}$ at layer $l$ and token position $t$, the SAE encoder produces sparse latent activations, and we denote the activation of feature $j$ by $a_j(x_{l,t})$.

Emotion control is implemented by selectively modulating the activation of emotion-related latent features, while leaving all other features unchanged:
\begin{equation}
a_j^{\text{new}}(x_{l,t}) =
\begin{cases}
a_j(x_{l,t}) + \alpha_e, & \text{if } j \in \mathcal{F}_e, \\
a_j(x_{l,t}), & \text{otherwise},
\end{cases}
\end{equation}
where $\alpha_e \in \mathbb{R}$ controls both the steering strength and direction. Positive values of $\alpha_e$ increase the activations of the selected emotion-related latent features to induce emotion $e$, whereas negative values suppress their activations to reduce the corresponding emotional expression. In our experiments, varying $\alpha_e$ provides a continuous control knob for emotion strength while keeping the selected latent-feature set fixed.

The decoder reconstructs the modified residual state:
\begin{equation}
x_{l,t}^{\text{new}} = b_{\text{pre}} + \sum_{j=1}^{n} a_j^{\text{new}}(x_{l,t})\, d_j.
\end{equation}
Under the linear reconstruction approximation, this corresponds to a direct residual stream intervention:
\begin{equation}
x_{l,t}^{\text{new}} \approx x_{l,t} + \alpha_e \sum_{j \in \mathcal{F}_e} d_j.
\end{equation}
This formulation makes explicit that emotion modulation operates as steering along a sparse combination of interpretable directions, rather than steering along a single dense global direction.

Because the same set of emotion-related latent features and a fixed scaling factor $\alpha_e$ are applied across all token positions, the control signal is time-invariant at the semantic-token level and does not introduce position-specific variation in the steering direction. The same formulation naturally allows time-varying variants in which $\alpha_e$ depends on token position.

\section{Experiments}
\subsection{Experimental Setup}
\paragraph{Location}
In LLM-based TTS architectures, SAE training can in principle be applied to different stages of the generation pipeline, including the autoregressive semantic backbone and downstream acoustic modules such as flow matching.
In this work, we train SAEs on residual stream activations from the semantic backbone during speech semantic-token generation.
This upstream location captures representations before acoustic synthesis, where linguistic content and high-level expressive variation are jointly organized.
It therefore allows us to analyze and modulate emotion-related factors before they are realized by the acoustic module, rather than directly modifying downstream acoustic dynamics.
To our knowledge, prior work has not explored SAE-based analysis of autoregressive semantic backbones in LLM-based TTS systems.
Following prior work applying SAEs to LLM residual streams \cite{lieberum2024gemma}, we adapt this residual-stream analysis to LLM-based TTS by training the SAE on residual stream activations from the IndexTTS2 semantic backbone \cite{zhou2025indextts2}.

\paragraph{SAE Training Dataset}
The SAE is trained on hidden representations extracted from 56,000 emotion-controlled TTS generations, evenly distributed across seven emotions (anger, disgust, fear, happiness, neutral, sadness, and surprise). The dataset consists of 400 distinct English texts, each synthesized under all combinations of seven emotions and 20 speaker-timbre references, yielding 2,800 utterances per timbre reference and 140 utterances per text.
Each generation specifies a target emotion, text content, an emotion-specific style reference, and a neutral reference utterance from IEMOCAP \cite{busso2008iemocap} used to fix speaker timbre.
The style reference provides emotional prosody, while the neutral reference fixes speaker identity.
This fully crossed design provides systematic coverage across emotions, texts, and speaker timbres, enabling emotion-related latent representations to be analyzed under matched text and speaker conditions.

\paragraph{SAE Training Details}
The SAE is trained on token-level representations extracted from the layer-16 pre-LayerNorm residual stream of the autoregressive semantic backbone.
To align SAE training with inference-time generation, we use only decode-phase hidden states, excluding conditioning and prefill activations that are not directly modified during steering.
These decode-phase representations constitute the SAE training data.
We employ a Top-$k$ sparse autoencoder with 4,096 latent features and $k=32$ active features per token.
The model is optimized using the reconstruction and auxiliary objectives described in Section~\ref{intro-sae}, with decoder columns constrained to unit norm.
Full implementation and training details are provided in Appendix~\ref{app:sae-training}.

\paragraph{Practicality and Computational Cost}
SAE training is a one-time offline step on frozen backbone activations.
In our setup, the SAE has 4,096 latent features and a 1280$\rightarrow$4096$\rightarrow$1280 architecture ($\approx$10.5M parameters), and is trained for 30,000 optimization steps on a single NVIDIA H100 GPU.
Most of the cost comes from extracting backbone activations, while SAE optimization itself is lightweight.
The trained SAE is small ($\approx$40MB in fp32), requires no updates to the TTS backbone, and supports inference-time feature interventions for both emotion induction and suppression.

\paragraph{Emotion Selectivity Analysis Dataset}
To compute emotion selectivity scores, we construct a separate analysis set of 43{,}408 emotion-controlled TTS generations, evenly distributed across four conditions: happiness, anger, sadness, and neutral.
Samples are generated with matched text and speaker-timbre references across emotion conditions, allowing latent activation statistics to be estimated under controlled text and speaker conditions.

\paragraph{Controlled Steering Evaluation Dataset}
\label{ana-dataset}
For controlled steering experiments, we construct a paired evaluation set for each target emotion, consisting of 100 matched text-speaker cases with one neutral sample and one target-emotion sample per case.
Each case is synthesized twice with the same text and speaker timbre: once using a neutral style reference and once using the target-emotion style reference.
This paired design enables direct measurement of emotion-induced latent modulation while controlling for text and speaker-dependent factors.

\paragraph{Evaluation Metrics}
Our evaluation covers three complementary levels. First, we assess SAE quality using reconstruction error and feature density analysis (Appendix~\ref{app:sae-eval}). Second, we test whether selected latent features have interpretable acoustic effects by measuring changes in mel-spectrogram structure, spectral centroid, and energy contours under controlled steering. Third, we evaluate downstream steering quality using three metrics: emotion similarity (Emo-SIM), Word Error Rate (WER), and speaker similarity (Spk-SIM).

Emo-SIM measures emotional consistency from emotion2vec embeddings~\cite{ma2024emotion2vec}, with emotion prototypes obtained by averaging reference utterances from IEMOCAP. WER measures content preservation and is computed with Whisper-Large V3~\cite{radford2023robust}. Spk-SIM measures speaker identity preservation using ERes2Net~\cite{chen2023enhanced}. Because speaker embeddings can also be affected by prosodic changes such as pitch, energy, and speaking rate, Spk-SIM should be interpreted as a conservative proxy for identity preservation under emotion conversion rather than a purely identity-only measure.

\begin{figure}[t]
  \begin{center}
    \centerline{\includegraphics[width=1\columnwidth]{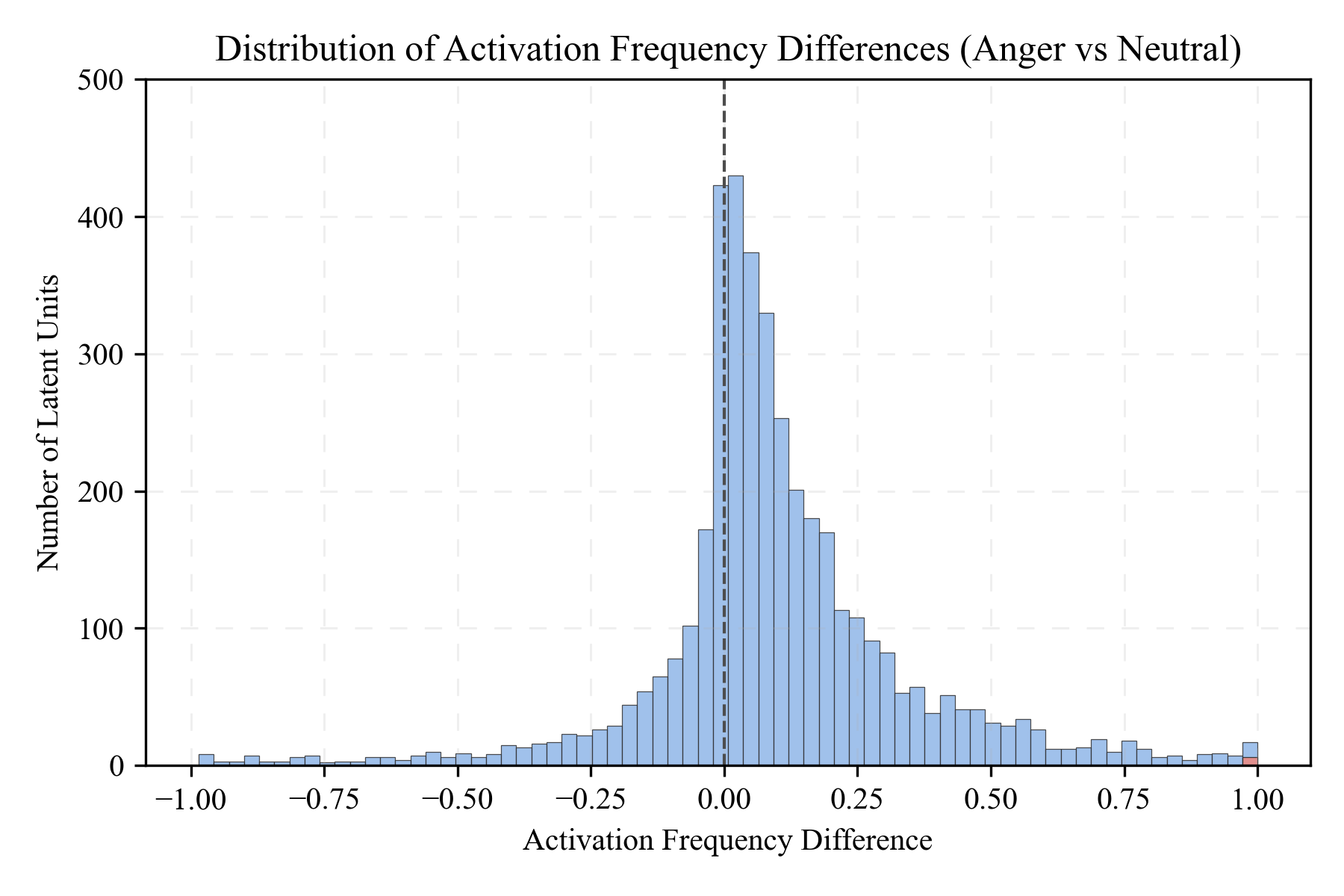}}
\caption{
Distribution of activation-frequency differences between anger and neutral conditions across all SAE latent features.
Each bar counts the number of latent features within a bin of the selectivity score $\Delta_i^{(e)}$.
The dashed vertical line marks zero difference; positive values indicate more frequent activation under anger than under the matched neutral condition.
}
    \label{fig:sae-emotion-selectivity-anger}
  \end{center}
  \vspace{-0.9cm}
\end{figure}
We further complement these metrics with human listening results using emotion and naturalness mean opinion scores (EMOS and NMOS) to assess perceived emotion accuracy and speech naturalness.

\subsection{Interpreting SAE Features}
\label{interpret}
We next investigate how emotion-related features are organized within the SAE latent space.
Rather than asking only whether emotion information is present in the backbone, we examine whether emotion-related modulation is broadly distributed across many latent features or concentrated in a small subset of sparse latent features.

\subsubsection{Sparse Organization of Emotion Signals}

To quantify how emotional modulation manifests in the latent space,
we compare each target emotion with its neutral counterpart under matched text and speaker conditions using the selectivity score defined in Equation~\ref{emo-score}.
Figure~\ref{fig:sae-emotion-selectivity-anger} shows the distribution for anger versus neutral.
The distribution is centered near zero, indicating that most latent features exhibit comparable activation frequencies across conditions.
Only a small subset displays substantial positive shifts, forming a sparse tail of emotion-related features.

For each target emotion, we rank latent features by their selectivity scores and use the top-$m$ features as candidates for emotion steering, with $m=6$ in our experiments.
For anger, each of the selected top-6 features achieves a selectivity score of 1 against neutral, indicating that they are activated under the target-emotion condition while remaining inactive under the matched neutral condition.
Although the SAE enforces token-level sparsity by construction, emotion selectivity is also concentrated at the dictionary level: the top-ranked features account for the strongest emotion-dependent activation differences.
Similar heavy-tailed patterns are observed for happiness and sadness (Appendix~\ref{app:emo-select}), suggesting that this sparse organization is consistent across emotions.

\subsubsection{Acoustic Validation of Selected Latent Features}
To assess whether selectivity-ranked latent features have interpretable acoustic effects, we intervene on individual latent features under matched text and speaker conditions, using neutral generation as the baseline.
\begin{figure}[t]
  \begin{center}
\centerline{\includegraphics[width=1.1\columnwidth, trim={0cm 0cm 0cm 1cm}, clip]{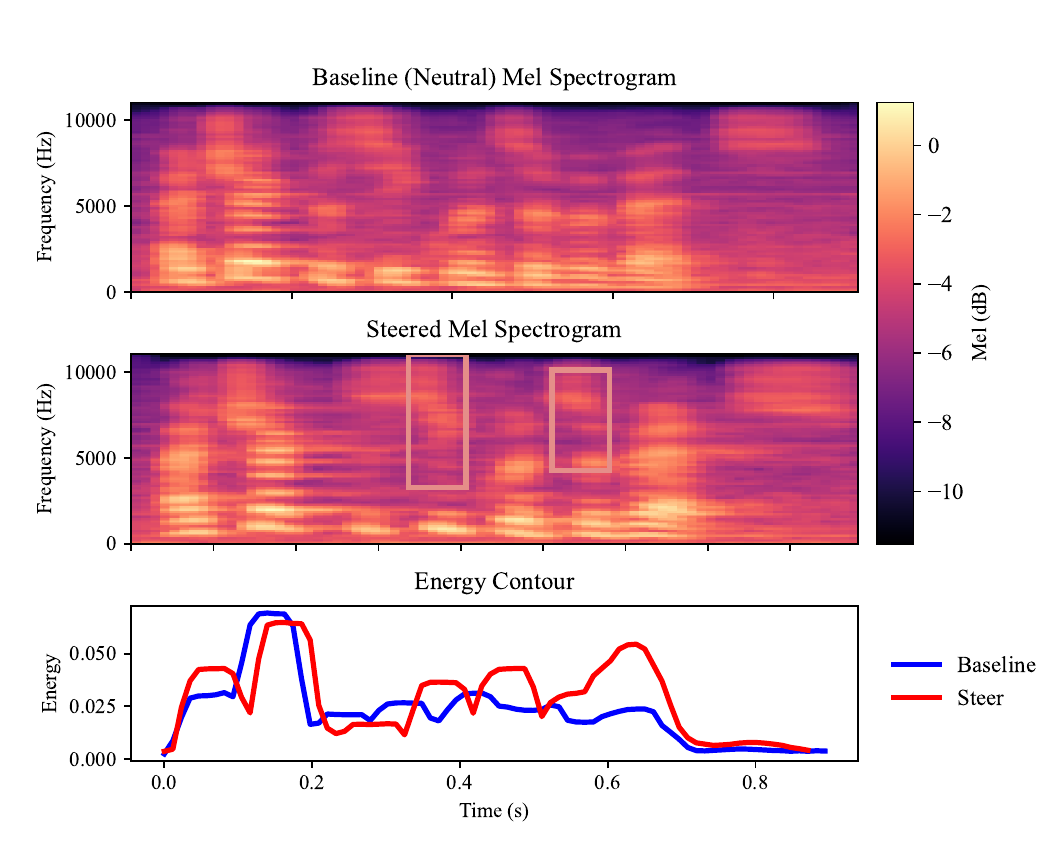}}
\caption{
Acoustic effects of steering one emotion-related latent feature (Latent Feature \#24).
Steering produces localized mid- to high-frequency amplification and local increases in short-time energy while largely preserving the overall time-frequency structure.
}

    \label{fig:mel}
  \end{center}
  \vspace{-0.7cm}
\end{figure}

\paragraph{Prosodic and Spectral Effects}
We examine mel-spectrogram representations and extract short-time energy (RMS) from the generated waveforms, discarding the first 5\% of frames to reduce onset artifacts.
Across evaluated utterances, positive steering consistently produces localized spectral amplification relative to the neutral baseline, most prominently in mid- to high-frequency regions.
As shown in Figure~\ref{fig:mel}, these spectral changes coincide with local increases in short-time energy, while the overall time-frequency structure remains largely preserved.

\paragraph{Quantitative Prosodic Analysis}
We further quantify the prosodic effects of the selected latent feature with a paired acoustic analysis under matched text and speaker conditions.
The intervention significantly increases mean F0 over voiced frames by $+23.11$ Hz ($p=1.07\times10^{-4}$) and RMS energy by $\Delta=+0.00435$ ($p=0.00769$), while effective duration shows no statistically significant change ($p=0.687$).
These results indicate that the selected latent feature modulates pitch and intensity without introducing a broad temporal change.
Full statistics and a paired F0 visualization are provided in Appendix~\ref{app:quantitative_acoustic_analysis}.

\paragraph{Spectral Structure (Brightness)}
We quantify spectral modulation induced by the selected latent feature by measuring changes in spectral centroid under controlled steering strengths.
Twenty neutral utterances are steered with symmetric scales from $-60$ to $+60$.
Spectral centroids are computed from magnitude STFT representations (16\,kHz sampling rate, 2048-point FFT, hop length 512), and frame-level values are averaged to obtain utterance-level centroids.
For each scale, centroid deviations are computed relative to the zero-scale condition and aggregated across utterances.

\begin{figure}[t]
  \begin{center}
    \centerline{\includegraphics[width=1\columnwidth]{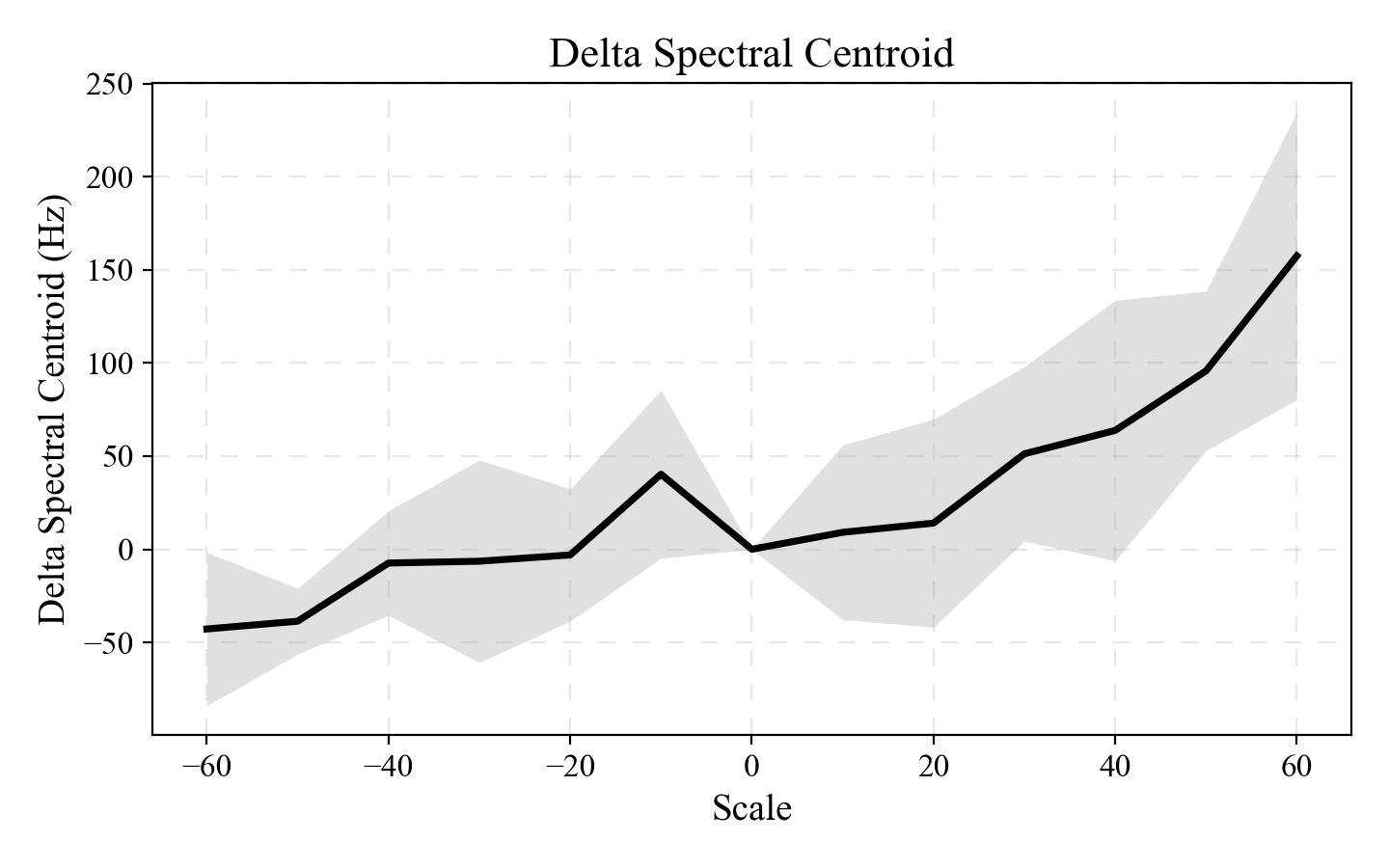}}
\caption{
Mean spectral-centroid deviation relative to the zero-scale baseline under steering scales from $-60$ to $+60$ across 20 neutral utterances.
Shaded regions denote 95\% confidence intervals.
Negative scales generally lower spectral centroid, whereas positive scales increase it, consistent with modulation of spectral brightness.
}
    \label{fig:centroid}
  \end{center}
  \vspace{-0.7cm}
\end{figure}

Figure~\ref{fig:centroid} reports mean centroid deviations with 95\% confidence intervals.
Centroid shifts vary systematically with steering strength: negative scales lower spectral centroid, whereas positive scales increase it.
This scale-dependent pattern indicates that the selected latent feature modulates spectral brightness.

\begin{table*}[t]
  \captionsetup{
    font=small,
    justification=raggedright,
    singlelinecheck=false,
  }
  \small
  \caption{
    Bidirectional emotion steering performance across three target emotions.
    Part I evaluates emotion induction (Neutral $\rightarrow$ Target), and Part II evaluates emotion suppression (Target $\rightarrow$ Neutral).
    Emo-SIM measures emotion similarity, WER measures content preservation, and Spk-SIM measures speaker similarity.
    Higher Emo-SIM and Spk-SIM are better; lower WER is better.
    Best results are bolded, and second-best results are underlined.
    }
  \centering
  \small
  \setlength{\tabcolsep}{4pt}
  \renewcommand{\arraystretch}{1.2}
  
  \resizebox{0.95\textwidth}{!}{
  \begin{tabular}{l ccc ccc ccc}
  \toprule
  & \multicolumn{3}{c}{\textbf{Anger}} 
  & \multicolumn{3}{c}{\textbf{Happiness}} 
  & \multicolumn{3}{c}{\textbf{Sadness}} \\
  \cmidrule(lr){2-4} \cmidrule(lr){5-7} \cmidrule(lr){8-10}
  Method 
  & Emo-SIM$\uparrow$ & WER$\downarrow$ & Spk-SIM$\uparrow$ 
  & Emo-SIM$\uparrow$ & WER$\downarrow$ & Spk-SIM$\uparrow$ 
  & Emo-SIM$\uparrow$ & WER$\downarrow$ & Spk-SIM$\uparrow$ \\
  
  \midrule
  \multicolumn{10}{c}{\textit{Part I: Emotion Induction (Neutral $\rightarrow$ Target)}} \\
  \midrule
  VALL-E-X \cite{zhang2023speak}
  & 0.831 & 3.1 &  0.302
  & 0.697 & 5.3 &  0.320
  & 0.869 & 7.8 &  0.352 \\
  
  Spark-TTS \cite{wang2025spark}
  & 0.857 & 2.7 &  0.488
  & 0.770 & 8.6 & 0.463 
  & \textbf{0.907} & 2.3 &  0.523 \\
  
  EmoVoice \cite{yang2025emovoice}
  & 0.806 & 4.1 & 0.358 
  & 0.728 & 3.4 & 0.342
  & 0.850 & 4.0 & 0.386 \\
  
  CosyVoice \cite{du2024cosyvoice1}
  & 0.813 & 3.9 & \underline{0.569}
  & 0.712 & \underline{2.9} & \textbf{0.597}
  & 0.799 & 2.4 & \underline{0.605} \\

  \midrule
  Random SAE ($m=6$)
  & 0.892 & 1.4 & \textbf{0.628}
  & 0.813 & 6.0 & 0.461
  & 0.858 & \underline{1.7} & \textbf{0.637} \\
  
  Global Steering
  & \underline{0.910} & \textbf{0.1} & 0.552
  & \underline{0.879} & 4.0 & 0.495
  & 0.876 & 1.9 & 0.516 \\
  
  SAE-Emotion (Ours)
  & \textbf{0.912} & \underline{0.3} & \underline{0.569}
  & \textbf{0.885} & \textbf{2.2} & \underline{0.515}
  & \underline{0.880} & \textbf{1.5} & 0.481 \\
  
  \midrule\midrule
  
  \multicolumn{10}{c}{\textit{Part II: Emotion Suppression (Target $\to$ Neutral)}} \\
  \midrule
  Random SAE ($m=6$)
  & 0.841 & \textbf{0.8} & 0.342 
  & 0.886 & \underline{2.14} & \underline{0.343} 
  & \underline{0.939} & \textbf{0.77} & 0.427 \\
  
  Global Steering
  & \underline{0.915} & \underline{2.6} & \textbf{0.392} 
  & \underline{0.920} & \textbf{1.48} & \textbf{0.379} 
  & 0.933 & 1.63 & \underline{0.436} \\
  
  SAE-Emotion (Ours)
  & \textbf{0.939} & 2.8 & \underline{0.374} 
  & \textbf{0.924} & 2.31 & 0.301 
  & \textbf{0.941} & \underline{0.80} & \textbf{0.441} \\
  
  \bottomrule
  \end{tabular}
  }
  \label{tab:bidirectional_steering_full}
  \end{table*}

\begin{figure}[t]
  \centering
  \includegraphics[width=\columnwidth]{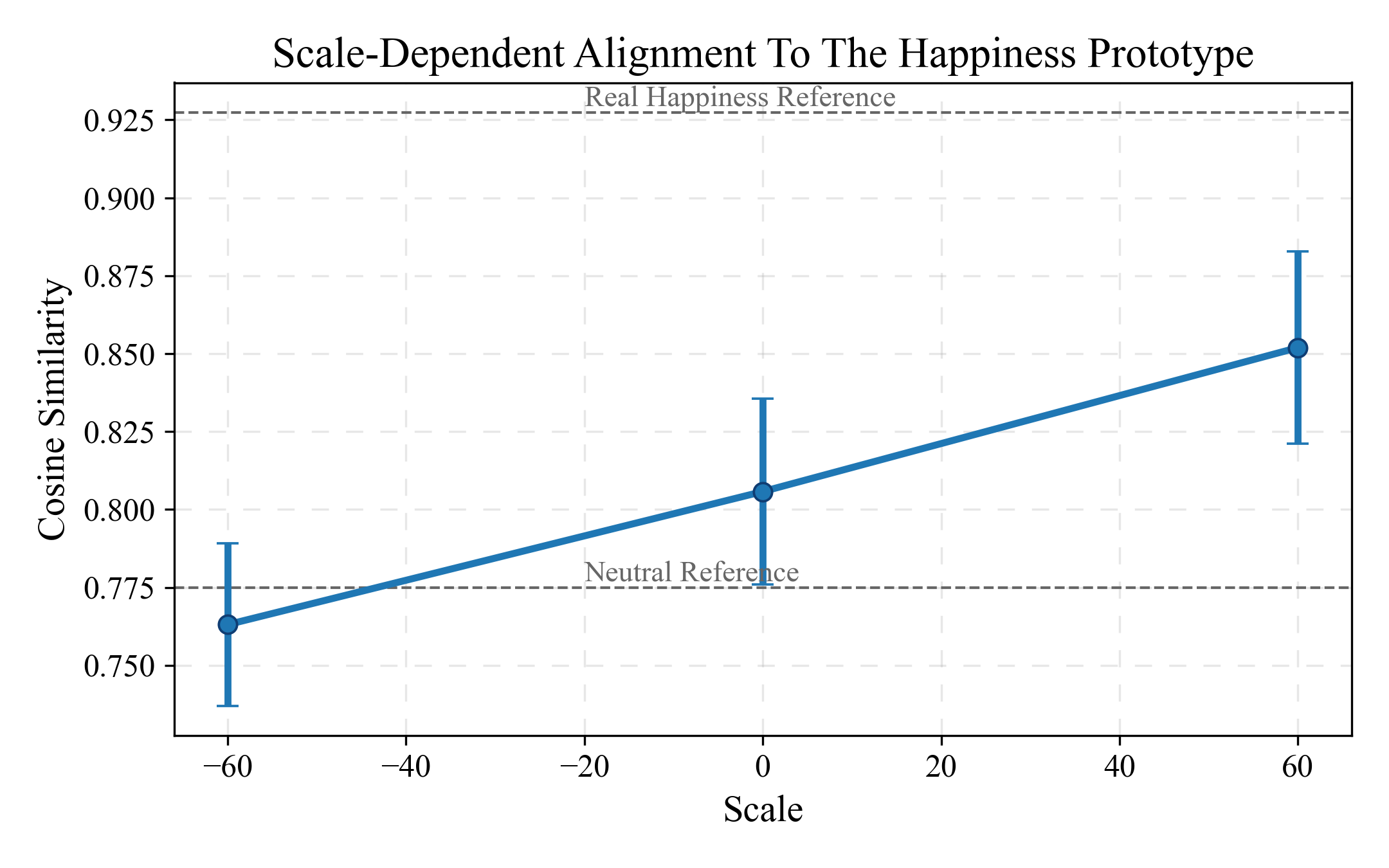}
 \caption{
Prototype-calibrated emotion-level alignment for the rank-1 happiness latent feature.
Cosine similarity is measured against an IEMOCAP-based happiness prototype; dashed lines mark real neutral and happy reference levels.
Error bars denote 95\% confidence intervals.
}
  \label{fig:calibrated-emotion}
  \vspace{-0.5cm}
\end{figure}

  \paragraph{Calibrated Emotion-Level Alignment}
  We further test whether steering a selected happiness-related latent feature moves generated speech toward a real happiness reference in emotion-embedding space.
  The latent feature is chosen as the rank-1 happiness feature under the sentence-level selectivity criterion, computed from paired happiness and neutral samples with matched text and speaker conditions.
  Using an IEMOCAP-based happiness prototype and real neutral and happiness reference levels, we apply scales $\{-60,0,+60\}$ to matched neutral text-speaker conditions.
  As shown in Figure~\ref{fig:calibrated-emotion}, increasing the steering scale moves generated samples toward the real happiness reference level, suggesting that the selected latent feature induces emotion-aligned variation rather than merely an uncalibrated acoustic shift.
  Additional dense scale sweeps and calibrated latent-feature budget analyses across emotions are provided in Appendix~\ref{app:emotion_selectivity_analyses}.

\subsection{Emotion Steering in TTS}
\paragraph{Steering Vector Construction from SAE Latent Features}

Individual latent features exhibit distinct acoustic characteristics, as shown in the preceding analysis.
However, emotional expression in speech is multifactorial, involving coordinated variation across multiple acoustic dimensions.
Steering along a single latent feature therefore captures only one aspect of emotion realization and may provide limited expressive flexibility as a generation control signal.

To reflect this multidimensional structure, we construct a composite steering direction by combining the top-6 emotion-related latent features ranked by their sentence-level selectivity scores.
We assign equal weights to these latent features and map the resulting combination back to the residual space using the corresponding SAE decoder directions, without including the decoder bias.

For comparison, we include two steering baselines.
\textit{Global steering} constructs a direction directly in the residual space using the mean-difference vector between target-emotion and neutral representations.
\textit{Random SAE steering} selects six latent features uniformly at random from the SAE latent space, combines them with equal weights, and maps the resulting combination to the residual space using the corresponding SAE decoder directions.
All steering vectors are applied using the formulation described in Section~\ref{steer}.

\paragraph{Steering Datasets and Compared Baselines}
We conduct emotion steering experiments on the datasets described in Section~\ref{ana-dataset}.
We compare SAE-Emotion with the steering baselines above and with existing TTS systems, including VALL-E-X, Spark-TTS, EmoVoice, and CosyVoice.
Detailed model descriptions, including the IndexTTS2 backbone used by our method, are provided in Appendix~\ref{baseline}, and evaluation protocols are provided in Appendix~\ref{baseline-test-detail}.

\paragraph{Bidirectional Steering Results}
Table~\ref{tab:bidirectional_steering_full} reports bidirectional emotion steering results for anger, happiness, and sadness.
We compare SAE-Emotion with existing TTS baselines and alternative steering methods under both induction and suppression settings.

In the induction setting, SAE-Emotion achieves the highest or second-highest Emo-SIM across all evaluated emotions while maintaining competitive WER.
Compared with Global Steering, SAE-Emotion uses a small set of emotion-related sparse latent features rather than a single dense mean-difference direction.
Compared with Random SAE, SAE-Emotion yields more consistent emotion alignment, suggesting that selectivity-based latent-feature selection is more reliable than arbitrary sparse perturbations.

In the suppression setting, SAE-Emotion achieves the highest or comparable Emo-SIM relative to the neutral reference across all three emotions, with WER and speaker similarity generally comparable to alternative steering methods.
These results indicate that SAE-Emotion supports both emotion induction and suppression without modifying the TTS backbone.
Additional analyses of cross-emotion behavior, latent-feature budget, and strong-steering robustness are provided in Appendices~\ref{app:cross_emotion_latent_budget} and~\ref{app:strong_steering}.

\begin{table}[t]
\centering
\small
\caption{
Human evaluation of SAE-Emotion, Global Steering, and Random SAE with 20 raters on a 0--5 scale.
EMOS measures perceived emotion accuracy, and NMOS measures naturalness; higher is better.
}
\label{tab:human_eval}
\begin{tabular}{lcc}
\toprule
Method & EMOS $\uparrow$ & NMOS $\uparrow$ \\
\midrule
SAE-Emotion & 3.22 & 3.49 \\
Global Steering & 3.10 & 3.38 \\
Random SAE & 1.82 & 3.22 \\
\bottomrule
\end{tabular}
\end{table}

\paragraph{Human Evaluation}
We further assess perceptual quality through a randomized blind listening study with 20 raters.
As shown in Table~\ref{tab:human_eval}, SAE-Emotion obtains the highest EMOS and NMOS among the compared steering methods.
The EMOS improvement indicates more accurate perceived expression of the target emotions, while the NMOS result suggests that the generated speech remains natural to listeners.
Together, these results support the perceptual quality of the induced emotional variations.

\section{Conclusion}
We investigated how emotion is internally represented in the semantic backbone of LLM-based TTS systems, showing that emotional variation is distributed across multiple sparse latent features.
Using sparse autoencoders to analyze semantic hidden states, we identified a small subset of emotion-related latent features with distinct activation and acoustic patterns under controlled conditions.
Building on these observations, we introduced a latent-feature intervention framework for bidirectional emotion control, enabling both emotion induction and suppression without modifying backbone parameters.
Our experiments show that manipulating selectivity-ranked sparse latent features can steer emotional expression while largely preserving linguistic content, naturalness, and speaker identity, providing an interpretable feature-level alternative to dense global steering.

\section*{Limitations}
Training sparse autoencoders on large-scale activation data introduces non-trivial computational and storage requirements.
Consequently, our full analysis is conducted on a single backbone configuration, with an additional cross-backbone analysis provided in the appendix.
Extending the framework to additional architectures with the same level of controlled evaluation would provide a more comprehensive assessment of generality.
In addition, emotional latent features in audio models are inherently difficult to quantify because speech expression is multidimensional and perceptual. In this work, we focus on representative and clearly observable latent features for controlled analysis. A more systematic characterization of the full latent space remains an open direction for future research.

\section*{Impact Statement}
This work proposes an interpretable framework for emotional steering in text-to-speech systems, which may benefit applications in accessibility, affective computing, and personalized speech synthesis.
At the same time, controllable emotional generation may be misused to produce emotionally manipulative or misleading speech, including impersonation or deceptive synthetic content.
Interpretability methods that reveal controllable internal factors may also be repurposed beyond the intended analysis setting.
We therefore encourage responsible deployment and continued research on safeguards for consent, misuse prevention, and detection of emotionally expressive synthetic speech.


\bibliography{1_example_paper}

@article{busso2008iemocap,
  title={{IEMOCAP}: Interactive emotional dyadic motion capture database},
  author={Busso, Carlos and Bulut, Murtaza and Lee, Chi-Chun and Kazemzadeh, Abe and Mower, Emily and Kim, Samuel and Chang, Jeannette N and Lee, Sungbok and Narayanan, Shrikanth S},
  journal={Language resources and evaluation},
  volume={42},
  number={4},
  pages={335--359},
  year={2008},
  publisher={Springer}
}

@article{bricken2023monosemanticity,
       title={Towards Monosemanticity: Decomposing Language Models With Dictionary Learning},
       author={Bricken, Trenton and Templeton, Adly and Batson, Joshua and Chen, Brian and Jermyn, Adam and Conerly, Tom and Turner, Nick and Anil, Cem and Denison, Carson and Askell, Amanda and Lasenby, Robert and Wu, Yifan and Kravec, Shauna and Schiefer, Nicholas and Maxwell, Tim and Joseph, Nicholas and Hatfield-Dodds, Zac and Tamkin, Alex and Nguyen, Karina and McLean, Brayden and Burke, Josiah E and Hume, Tristan and Carter, Shan and Henighan, Tom and Olah, Christopher},
       year={2023},
       journal={Transformer Circuits Thread},
       note={https://transformer-circuits.pub/2023/monosemantic-features/index.html}
    }

@inproceedings{yun2021transformer,
  title={Transformer visualization via dictionary learning: contextualized embedding as a linear superposition of transformer factors},
  author={Yun, Zeyu and Chen, Yubei and Olshausen, Bruno and LeCun, Yann},
  booktitle={Proceedings of Deep Learning Inside Out (DeeLIO): The 2nd Workshop on Knowledge Extraction and Integration for Deep Learning Architectures},
  pages={1--10},
  year={2021}
}

@article{elhage2022toy,
  title={Toy models of superposition},
  author={Elhage, Nelson and Hume, Tristan and Olsson, Catherine and Schiefer, Nicholas and Henighan, Tom and Kravec, Shauna and Hatfield-Dodds, Zac and Lasenby, Robert and Drain, Dawn and Chen, Carol and others},
  journal={arXiv preprint arXiv:2209.10652},
  year={2022}
}

@article{cunningham2023sparse,
  title={Sparse autoencoders find highly interpretable features in language models},
  author={Cunningham, Hoagy and Ewart, Aidan and Riggs, Logan and Huben, Robert and Sharkey, Lee},
  journal={arXiv preprint arXiv:2309.08600},
  year={2023}
}

@misc{durmusevaluating,
author = {Esin Durmus and Alex Tamkin and Jack Clark and Jerry Wei and Jonathan Marcus and Joshua Batson and Kunal Handa and Liane Lovitt and Meg Tong and Miles McCain and Oliver Rausch and Saffron Huang and Sam Bowman and Stuart Ritchie and Tom Henighan and Deep Ganguli},
title = {Evaluating Feature Steering: A Case Study in Mitigating Social Biases},
date = {2024-10-25},
year = {2024},
url = {https://anthropic.com/research/evaluating-feature-steering},
}

@article{templeton2024scaling,
       title={Scaling Monosemanticity: Extracting Interpretable Features from Claude 3 Sonnet},
       author={Templeton, Adly and Conerly, Tom and Marcus, Jonathan and Lindsey, Jack and Bricken, Trenton and Chen, Brian and Pearce, Adam and Citro, Craig and Ameisen, Emmanuel and Jones, Andy and Cunningham, Hoagy and Turner, Nicholas L and McDougall, Callum and MacDiarmid, Monte and Freeman, C. Daniel and Sumers, Theodore R. and Rees, Edward and Batson, Joshua and Jermyn, Adam and Carter, Shan and Olah, Chris and Henighan, Tom},
       year={2024},
       journal={Transformer Circuits Thread},
       url={https://transformer-circuits.pub/2024/scaling-monosemanticity/index.html}
    }

@article{lou2025sae,
  title={Sae-v: Interpreting multimodal models for enhanced alignment},
  author={Lou, Hantao and Li, Changye and Ji, Jiaming and Yang, Yaodong},
  journal={arXiv preprint arXiv:2502.17514},
  year={2025}
}

@article{joseph2025steering,
  title={Steering {CLIP}'s vision transformer with sparse autoencoders},
  author={Joseph, Sonia and Suresh, Praneet and Goldfarb, Ethan and Hufe, Lorenz and Gandelsman, Yossi and Graham, Robert and Bzdok, Danilo and Samek, Wojciech and Richards, Blake Aaron},
  journal={arXiv preprint arXiv:2504.08729},
  year={2025}
}

@article{cywinski2025saeuron,
  title={{SAeUron}: Interpretable concept unlearning in diffusion models with sparse autoencoders},
  author={Cywi{\'n}ski, Bartosz and Deja, Kamil},
  journal={arXiv preprint arXiv:2501.18052},
  year={2025}
}

@article{kim2025concept,
  title={Concept steerers: Leveraging k-sparse autoencoders for controllable generations},
  author={Kim, Dahye and Ghadiyaram, Deepti},
  journal={arXiv preprint arXiv:2501.19066},
  year={2025}
}

@article{paek2025learning,
  title={Learning Interpretable Features in Audio Latent Spaces via Sparse Autoencoders},
  author={Paek, Nathan and Zang, Yongyi and Yang, Qihui and Leistikow, Randal},
  journal={arXiv preprint arXiv:2510.23802},
  year={2025}
}

@article{ferrando2024know,
  title={Do {I} know this entity? knowledge awareness and hallucinations in language models},
  author={Ferrando, Javier and Obeso, Oscar and Rajamanoharan, Senthooran and Nanda, Neel},
  journal={arXiv preprint arXiv:2411.14257},
  year={2024}
}

@inproceedings{wang2018style,
  title={Style tokens: Unsupervised style modeling, control and transfer in end-to-end speech synthesis},
  author={Wang, Yuxuan and Stanton, Daisy and Zhang, Yu and Ryan, RJ-Skerry and Battenberg, Eric and Shor, Joel and Xiao, Ying and Jia, Ye and Ren, Fei and Saurous, Rif A},
  booktitle={International conference on machine learning},
  pages={5180--5189},
  year={2018},
  organization={PMLR}
}

@inproceedings{radford2023robust,
  title={Robust speech recognition via large-scale weak supervision},
  author={Radford, Alec and Kim, Jong Wook and Xu, Tao and Brockman, Greg and McLeavey, Christine and Sutskever, Ilya},
  booktitle={International conference on machine learning},
  pages={28492--28518},
  year={2023},
  organization={PMLR}
}

@inproceedings{ma2024emotion2vec,
  title={emotion2vec: Self-supervised pre-training for speech emotion representation},
  author={Ma, Ziyang and Zheng, Zhisheng and Ye, Jiaxin and Li, Jinchao and Gao, Zhifu and Zhang, Shiliang and Chen, Xie},
  booktitle={Findings of the Association for Computational Linguistics: ACL 2024},
  pages={15747--15760},
  year={2024}
}

@article{li2023inference,
  title={Inference-time intervention: Eliciting truthful answers from a language model},
  author={Li, Kenneth and Patel, Oam and Vi{\'e}gas, Fernanda and Pfister, Hanspeter and Wattenberg, Martin},
  journal={Advances in Neural Information Processing Systems},
  volume={36},
  pages={41451--41530},
  year={2023}
}

@article{liu2023context,
  title={In-context vectors: Making in context learning more effective and controllable through latent space steering},
  author={Liu, Sheng and Ye, Haotian and Xing, Lei and Zou, James},
  journal={arXiv preprint arXiv:2311.06668},
  year={2023}
}

@article{subramani2022extracting,
  title={Extracting latent steering vectors from pretrained language models},
  author={Subramani, Nishant and Suresh, Nivedita and Peters, Matthew E},
  journal={arXiv preprint arXiv:2205.05124},
  year={2022}
}

@inproceedings{wang2025adaptive,
  title={Adaptive activation steering: A tuning-free llm truthfulness improvement method for diverse hallucinations categories},
  author={Wang, Tianlong and Jiao, Xianfeng and Zhu, Yinghao and Chen, Zhongzhi and He, Yifan and Chu, Xu and Gao, Junyi and Wang, Yasha and Ma, Liantao},
  booktitle={Proceedings of the ACM on Web Conference 2025},
  pages={2562--2578},
  year={2025}
}

@article{gavrikov2024can,
  title={Can we talk models into seeing the world differently?},
  author={Gavrikov, Paul and Lukasik, Jovita and Jung, Steffen and Geirhos, Robert and Mirza, M Jehanzeb and Keuper, Margret and Keuper, Janis},
  journal={arXiv preprint arXiv:2403.09193},
  year={2024}
}

@article{li2024get,
  title={Get what you want, not what you don't: Image content suppression for text-to-image diffusion models},
  author={Li, Senmao and van de Weijer, Joost and Hu, Taihang and Khan, Fahad Shahbaz and Hou, Qibin and Wang, Yaxing and Yang, Jian},
  journal={arXiv preprint arXiv:2402.05375},
  year={2024}
}

@article{zhou2025indextts2,
  title={{IndexTTS2}: A Breakthrough in Emotionally Expressive and Duration-Controlled Auto-Regressive Zero-Shot Text-to-Speech},
  author={Zhou, Siyi and Zhou, Yiquan and He, Yi and Zhou, Xun and Wang, Jinchao and Deng, Wei and Shu, Jingchen},
  journal={arXiv preprint arXiv:2506.21619},
  year={2025}
}

@article{deng2025indextts,
  title={{IndexTTS}: An industrial-level controllable and efficient zero-shot text-to-speech system},
  author={Deng, Wei and Zhou, Siyi and Shu, Jingchen and Wang, Jinchao and Wang, Lu},
  journal={arXiv preprint arXiv:2502.05512},
  year={2025}
}

@article{du2024cosyvoice1,
  title={{CosyVoice}: A scalable multilingual zero-shot text-to-speech synthesizer based on supervised semantic tokens},
  author={Du, Zhihao and Chen, Qian and Zhang, Shiliang and Hu, Kai and Lu, Heng and Yang, Yexin and Hu, Hangrui and Zheng, Siqi and Gu, Yue and Ma, Ziyang and others},
  journal={arXiv preprint arXiv:2407.05407},
  year={2024}
}

@article{du2024cosyvoice2,
  title={{CosyVoice 2}: Scalable streaming speech synthesis with large language models},
  author={Du, Zhihao and Wang, Yuxuan and Chen, Qian and Shi, Xian and Lv, Xiang and Zhao, Tianyu and Gao, Zhifu and Yang, Yexin and Gao, Changfeng and Wang, Hui and others},
  journal={arXiv preprint arXiv:2412.10117},
  year={2024}
}

@article{xie2025emosteer,
  title={{EmoSteer-TTS}: Fine-Grained and Training-Free Emotion-Controllable Text-to-Speech via Activation Steering},
  author={Xie, Tianxin and Yang, Shan and Li, Chenxing and Yu, Dong and Liu, Li},
  journal={arXiv preprint arXiv:2508.03543},
  year={2025}
}

@article{du2025cosyvoice3,
  title={{CosyVoice 3}: Towards in-the-wild speech generation via scaling-up and post-training},
  author={Du, Zhihao and Gao, Changfeng and Wang, Yuxuan and Yu, Fan and Zhao, Tianyu and Wang, Hao and Lv, Xiang and Wang, Hui and Ni, Chongjia and Shi, Xian and others},
  journal={arXiv preprint arXiv:2505.17589},
  year={2025}
}

@article{zhang2023speak,
  title={Speak foreign languages with your own voice: Cross-lingual neural codec language modeling},
  author={Zhang, Ziqiang and Zhou, Long and Wang, Chengyi and Chen, Sanyuan and Wu, Yu and Liu, Shujie and Chen, Zhuo and Liu, Yanqing and Wang, Huaming and Li, Jinyu and others},
  journal={arXiv preprint arXiv:2303.03926},
  year={2023}
}

@article{wang2025spark,
  title={{Spark-TTS}: An efficient {LLM}-based text-to-speech model with single-stream decoupled speech tokens},
  author={Wang, Xinsheng and Jiang, Mingqi and Ma, Ziyang and Zhang, Ziyu and Liu, Songxiang and Li, Linqin and Liang, Zheng and Zheng, Qixi and Wang, Rui and Feng, Xiaoqin and others},
  journal={arXiv preprint arXiv:2503.01710},
  year={2025}
}

@article{guo2024text,
  title={Text-aware and Context-aware Expressive Audiobook Speech Synthesis},
  author={Guo, Dake and Zhu, Xinfa and Xue, Liumeng and Zhang, Yongmao and Tian, Wenjie and Xie, Lei},
  journal={arXiv preprint arXiv:2406.05672},
  year={2024}
}

@inproceedings{wadley2022future,
  title={The future of emotion in human-computer interaction},
  author={Wadley, Greg and Kostakos, Vassilis and Koval, Peter and Smith, Wally and Webber, Sarah and Cox, Anna and Gross, James J and H{\"o}{\"o}k, Kristina and Mandryk, Regan and Slov{\'a}k, Petr},
  booktitle={CHI Conference on human factors in computing systems extended abstracts},
  pages={1--6},
  year={2022}
}

@article{kim2021expressive,
  title={Expressive text-to-speech using style tag},
  author={Kim, Minchan and Cheon, Sung Jun and Choi, Byoung Jin and Kim, Jong Jin and Kim, Nam Soo},
  journal={arXiv preprint arXiv:2104.00436},
  year={2021}
}

@inproceedings{yang2025emovoice,
  title={{EmoVoice}: {LLM}-based emotional text-to-speech model with freestyle text prompting},
  author={Yang, Guanrou and Yang, Chen and Chen, Qian and Ma, Ziyang and Chen, Wenxi and Wang, Wen and Wang, Tianrui and Yang, Yifan and Niu, Zhikang and Liu, Wenrui and others},
  booktitle={Proceedings of the 33rd ACM International Conference on Multimedia},
  pages={10748--10757},
  year={2025}
}

@inproceedings{ji2024textrolspeech,
  title={{TextrolSpeech}: A text style control speech corpus with codec language text-to-speech models},
  author={Ji, Shengpeng and Zuo, Jialong and Fang, Minghui and Jiang, Ziyue and Chen, Feiyang and Duan, Xinyu and Huai, Baoxing and Zhao, Zhou},
  booktitle={ICASSP 2024-2024 IEEE International Conference on Acoustics, Speech and Signal Processing (ICASSP)},
  pages={10301--10305},
  year={2024},
  organization={IEEE}
}

@inproceedings{ji2025controlspeech,
    title = "{C}ontrol{S}peech: Towards Simultaneous and Independent Zero-shot Speaker Cloning and Zero-shot Language Style Control",
    author = "Ji, Shengpeng  and
      Chen, Qian  and
      Wang, Wen  and
      Zuo, Jialong  and
      Fang, Minghui  and
      Jiang, Ziyue  and
      Huang, Hai  and
      Wang, Zehan  and
      Cheng, Xize  and
      Zheng, Siqi  and
      Zhao, Zhou",
    editor = "Che, Wanxiang  and
      Nabende, Joyce  and
      Shutova, Ekaterina  and
      Pilehvar, Mohammad Taher",
    booktitle = "Proceedings of the 63rd Annual Meeting of the Association for Computational Linguistics (Volume 1: Long Papers)",
    month = jul,
    year = "2025",
    address = "Vienna, Austria",
    publisher = "Association for Computational Linguistics",
    url = "https://aclanthology.org/2025.acl-long.346/",
    doi = "10.18653/v1/2025.acl-long.346",
    pages = "6966--6981",
    ISBN = "979-8-89176-251-0"
}

@article{liu2025uddetts,
  title={{UDDETTS}: Unifying Discrete and Dimensional Emotions for Controllable Emotional Text-to-Speech},
  author={Liu, Jiaxuan and Xiang, Yang and Zhao, Han and Li, Xiangang and Gao, Yingying and Zhang, Shilei and Ling, Zhenhua},
  journal={arXiv preprint arXiv:2505.10599},
  year={2025}
}

@article{olshausen1997sparse,
  title={Sparse coding with an overcomplete basis set: A strategy employed by V1?},
  author={Olshausen, Bruno A and Field, David J},
  journal={Vision research},
  volume={37},
  number={23},
  pages={3311--3325},
  year={1997},
  publisher={Elsevier}
}

@article{lee2017emotional,
  title={Emotional end-to-end neural speech synthesizer},
  author={Lee, Younggun and Rabiee, Azam and Lee, Soo-Young},
  journal={arXiv preprint arXiv:1711.05447},
  year={2017}
}

@inproceedings{skerry2018towards,
  title={Towards end-to-end prosody transfer for expressive speech synthesis with tacotron},
  author={Skerry-Ryan, RJ and Battenberg, Eric and Xiao, Ying and Wang, Yuxuan and Stanton, Daisy and Shor, Joel and Weiss, Ron and Clark, Rob and Saurous, Rif A},
  booktitle={international conference on machine learning},
  pages={4693--4702},
  year={2018},
  organization={PMLR}
}

@article{jia2018transfer,
  title={Transfer learning from speaker verification to multispeaker text-to-speech synthesis},
  author={Jia, Ye and Zhang, Yu and Weiss, Ron and Wang, Quan and Shen, Jonathan and Ren, Fei and Nguyen, Patrick and Pang, Ruoming and Lopez Moreno, Ignacio and Wu, Yonghui and others},
  journal={Advances in neural information processing systems},
  volume={31},
  year={2018}
}

@article{gao2024scaling,
  title={Scaling and evaluating sparse autoencoders},
  author={Gao, Leo and la Tour, Tom Dupr{\'e} and Tillman, Henk and Goh, Gabriel and Troll, Rajan and Radford, Alec and Sutskever, Ilya and Leike, Jan and Wu, Jeffrey},
  journal={arXiv preprint arXiv:2406.04093},
  year={2024}
}

@article{makhzani2013k,
  title={K-sparse autoencoders},
  author={Makhzani, Alireza and Frey, Brendan},
  journal={arXiv preprint arXiv:1312.5663},
  year={2013}
}

@article{lieberum2024gemma,
  title={Gemma scope: Open sparse autoencoders everywhere all at once on gemma 2},
  author={Lieberum, Tom and Rajamanoharan, Senthooran and Conmy, Arthur and Smith, Lewis and Sonnerat, Nicolas and Varma, Vikrant and Kram{\'a}r, J{\'a}nos and Dragan, Anca and Shah, Rohin and Nanda, Neel},
  journal={arXiv preprint arXiv:2408.05147},
  year={2024}
}

@article{kingma2014adam,
  title={Adam: A method for stochastic optimization},
  author={Kingma, Diederik P and Ba, Jimmy},
  journal={arXiv preprint arXiv:1412.6980},
  year={2014}
}

@article{chen2023enhanced,
  title={An enhanced {Res2Net} with local and global feature fusion for speaker verification},
  author={Chen, Yafeng and Zheng, Siqi and Wang, Hui and Cheng, Luyao and Chen, Qian and Qi, Jiajun},
  journal={arXiv preprint arXiv:2305.12838},
  year={2023}
}

@article{wang2023neural,
  title={Neural codec language models are zero-shot text to speech synthesizers},
  author={Wang, Chengyi and Chen, Sanyuan and Wu, Yu and Zhang, Ziqiang and Zhou, Long and Liu, Shujie and Chen, Zhuo and Liu, Yanqing and Wang, Huaming and Li, Jinyu and others},
  journal={arXiv preprint arXiv:2301.02111},
  year={2023}
}

@article{ye2025llasa,
  title={{LLaSA}: Scaling train-time and inference-time compute for {LLaMA}-based speech synthesis},
  author={Ye, Zhen and Zhu, Xinfa and Chan, Chi-Min and Wang, Xinsheng and Tan, Xu and Lei, Jiahe and Peng, Yi and Liu, Haohe and Jin, Yizhu and Dai, Zheqi and others},
  journal={arXiv preprint arXiv:2502.04128},
  year={2025}
}
\bibliographystyle{icml2026} 

\newpage
\appendix
\onecolumn
\captionsetup{font=small}
\setlength{\textfloatsep}{10pt plus 2pt minus 2pt}
\setlength{\floatsep}{8pt plus 2pt minus 2pt}
\setlength{\intextsep}{8pt plus 2pt minus 2pt}

\section*{Appendix Organization}
The appendix follows the order of the main paper narrative. Appendix~\ref{app:text-prompt} describes the controlled prompts and metadata format. Appendices~\ref{app:sae-training}--\ref{app:sae-eval} report SAE training and intrinsic evaluation details. Appendices~\ref{app:emotion_selectivity_analyses}--\ref{app:strong_steering} provide additional analyses of emotion selectivity, latent-feature control, acoustic effects, and strong steering. Appendices~\ref{baseline}--\ref{baseline-test-detail} describe baselines and evaluation protocols. Appendix~\ref{app:llasa_analysis} provides an additional cross-backbone analysis.

\section{Text Prompts Used for Controlled Emotion Analysis}
\label{app:text-prompt}
\paragraph{Prompt Design}
To isolate emotion-specific modulation from lexical content, 
we construct a set of 100 semantically neutral English sentences. 
The prompts describe everyday events, procedural actions, 
or observational statements without explicit affective language. 
This design minimizes intrinsic emotional bias in the text 
and ensures that differences in generated speech arise 
primarily from emotional style conditioning.

\paragraph{Controlled Generation Protocol}
For each text prompt, we generate speech under multiple emotional reference conditions 
(e.g., neutral, happiness, sadness, anger). 
We fix the speaker identity and timbre reference across emotion conditions, 
and vary only the emotional style reference. 
This controlled setup ensures that any differences in hidden activations 
can be attributed to emotional style rather than lexical content or speaker-dependent variation.

\paragraph{Metadata Format}
Each sample is stored as a structured JSON record 
containing the emotion label, text prompt, style reference, 
timbre reference, and speaker identifier. 
An example entry is shown below:
\begin{verbatim}
{
  "emotion": "Anger",
  "text": "I arrived at the location earlier than expected.",
  "style_ref": "Anger_Ses03M_impro01_F025.wav",
  "timbre_ref": "Ses01F_impro04_M016.wav",
  "speaker_id": "0001",
  "text_id": 0,
  "utterance_id": 0
}
\end{verbatim}

\paragraph{Full Prompt List}
The complete set of 100 neutral prompts used in controlled emotion analysis 
is listed below.

\begingroup
\small
\setlength{\columnsep}{0.25in}
\begin{multicols}{2}
\begin{enumerate}
\setlength{\itemsep}{0pt}
\setlength{\parskip}{0pt}
\item I arrived at the location earlier than expected.
\item The meeting started a few minutes late.
\item I noticed a change in the schedule today.
\item The room was quieter than usual.
\item I reviewed the document before sending it.
\item The package was delivered this morning.
\item I checked the results after the test finished.
\item The system updated overnight.
\item I received the message earlier today.
\item The lights were turned off automatically.
\item I waited for the response before continuing.
\item The train stopped at the next station.
\item I followed the instructions as written.
\item The report includes several data points.
\item I observed a difference in the output.
\item The process completed without interruption.
\item I took notes during the discussion.
\item The file was saved in the correct folder.
\item I compared the two versions carefully.
\item The door remained open for a while.
\item I checked the time before leaving.
\item The equipment was moved to another room.
\item I reviewed the details again later.
\item The announcement was made earlier today.
\item I noticed the change after returning.
\item The temperature stayed the same throughout the day.
\item I updated the record after verification.
\item The call ended shortly after noon.
\item I followed up on the request yesterday.
\item The task required several steps to complete.
\item I looked over the information provided.
\item The session lasted longer than planned.
\item I adjusted the settings accordingly.
\item The message appeared on the screen.
\item I observed the process from start to finish.
\item The document was shared with the team.
\item I recorded the results for later use.
\item The system responded as expected.
\item I confirmed the details this afternoon.
\item The data was collected over several days.
\item I noticed the update after restarting the system.
\item The device remained connected throughout the test.
\item I reviewed the notes before the meeting.
\item The instructions were displayed clearly.
\item I checked the list once more.
\item The session resumed after a short break.
\item I observed a pattern in the results.
\item The file was uploaded successfully.
\item I followed the procedure step by step.
\item The results were stored securely.
\item I verified the information before proceeding.
\item The response arrived later in the day.
\item I adjusted the timeline slightly.
\item The screen displayed the final output.
\item I reviewed the changes carefully.
\item The update affected several components.
\item I noted the difference during testing.
\item The system remained stable throughout.
\item I checked the configuration settings.
\item The process required additional input.
\item I observed the behavior under different conditions.
\item The report was completed on time.
\item I reviewed the summary afterward.
\item The message was forwarded to the team.
\item I confirmed the location before arriving.
\item The system logged the activity automatically.
\item I checked the connection status.
\item The results matched the initial expectations.
\item I reviewed the timeline again.
\item The device responded to the input.
\item I noted the changes in the display.
\item The update was applied successfully.
\item I checked the documentation for details.
\item The system processed the request.
\item I observed the output during the test.
\item The file was accessed earlier.
\item I reviewed the configuration afterward.
\item The process continued without delay.
\item I checked the status before leaving.
\item The message was received as scheduled.
\item I observed the system behavior overnight.
\item The document was revised slightly.
\item I reviewed the output once more.
\item The update was completed earlier.
\item I checked the log for details.
\item The process required confirmation.
\item I noted the response time.
\item The system remained active.
\item I reviewed the settings again.
\item The task was completed as planned.
\item I observed the results after processing.
\item The data was organized by category.
\item I checked the sequence of events.
\item The update included minor changes.
\item I reviewed the procedure carefully.
\item The system recorded the input.
\item I noted the final output.
\item The file was closed properly.
\item I checked the summary afterward.
\item The process concluded successfully.
\end{enumerate}
\end{multicols}
\endgroup

\section{SAE Training Details}
\label{app:sae-training}

\paragraph{Optimization Setup.}
We train the SAE for 30,000 optimization steps using Adam \cite{kingma2014adam} 
($\mathrm{lr}=10^{-4}$, $\epsilon=6.25\times10^{-16}$), 
processing a target of 16,384 tokens per update. 
The latent dimensionality is fixed to 4096, with $K=32$ active features enforced per token throughout training.

\paragraph{Loss and Regularization.}
Training minimizes a normalized reconstruction objective. 
To reduce feature collapse and mitigate dead latent features, 
we incorporate a usage-based sparsity regularizer (weight 0.01) 
together with an auxiliary residual projection loss ($\lambda_{\text{aux}}=0.1$). 
A latent feature is considered dead if it has not been activated within the last $10^6$ tokens.

\paragraph{Stability Mechanisms.}
Decoder columns are renormalized to unit norm after each update to maintain consistent intervention scales. 
Decoder gradients are additionally projected onto the orthogonal subspace of the weight vectors to preserve the norm constraint. 
We further apply exponential moving average (EMA, decay 0.99) to stabilize optimization.

\paragraph{Latent Dimension Sensitivity.}
We also trained a 10,240-dimensional SAE. 
Although reconstruction error further decreased, 
emotion-related features became more fragmented and distributed across a larger number of sparse directions, 
reducing emotional interpretability in the TTS residual space. 
For clarity and controlled analysis of emotion steering, 
we therefore focus on the 4096-dimensional model.



\paragraph{Training Infrastructure.}
All experiments are conducted on a single NVIDIA H100 GPU.

\section{SAE Evaluation}
\label{app:sae-eval}
We evaluate the sparsity and reconstruction behavior of the trained SAE through feature density and reconstruction error analysis. The trained SAE achieves a normalized mean squared reconstruction error (MSE) of 0.129 on centered and unit-normalized residual representations, indicating that the sparse latent activations preserve most of the semantic variance despite the strict Top-$k$ constraint.

\begin{figure}[htbp]
  \begin{center}
    \centerline{\includegraphics[width=0.55\textwidth]{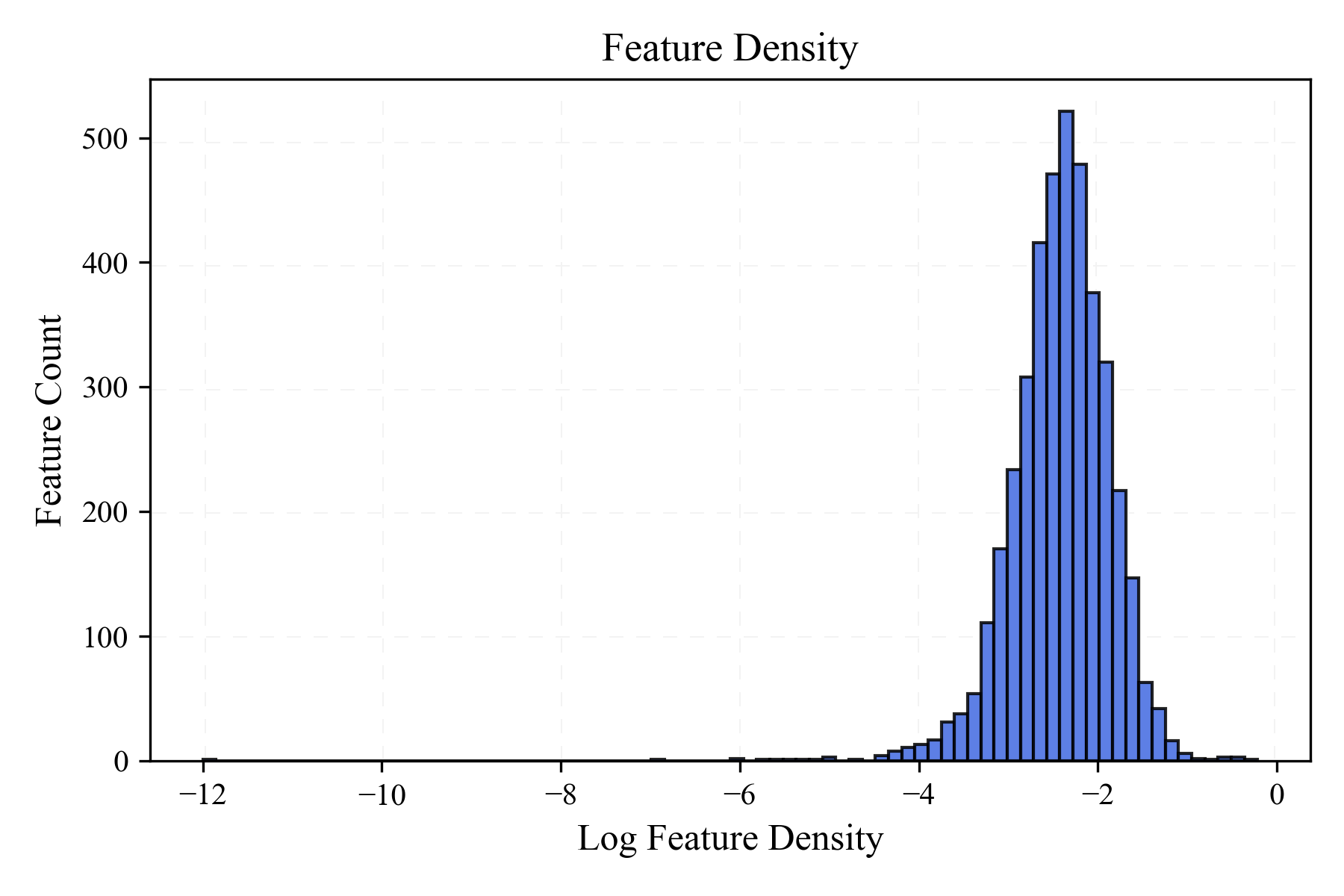}}
    \caption{
      Feature activation density distribution on a logarithmic scale. The distribution is long-tailed but does not contain a large mass of inactive latent features, indicating stable Top-$k$ sparse activations.
    }
    \label{app:density}
  \end{center}
\end{figure}

Figure~\ref{app:density} shows the distribution of feature activation densities on a logarithmic scale. The average activation fraction per token matches the theoretical Top-$k$ sparsity level $(32/4096 \approx 0.0078)$, confirming proper sparsity enforcement. The density distribution exhibits a smooth long-tailed pattern without a large mass near zero, and no dead latent features are observed under the $10^6$-token inactivity criterion. These results indicate stable and well-utilized sparse feature representations.

\section{Additional Emotion Selectivity and Control Analyses}

\label{app:emotion_selectivity_analyses}
This section provides supplementary analyses of emotion selectivity, intensity control, cross-emotion behavior, latent-feature overlap, and latent-feature selection criteria.

\begin{figure}[htbp]
  \begin{center}
    \begin{subfigure}{0.48\columnwidth}
      \centering
      \includegraphics[width=\linewidth]{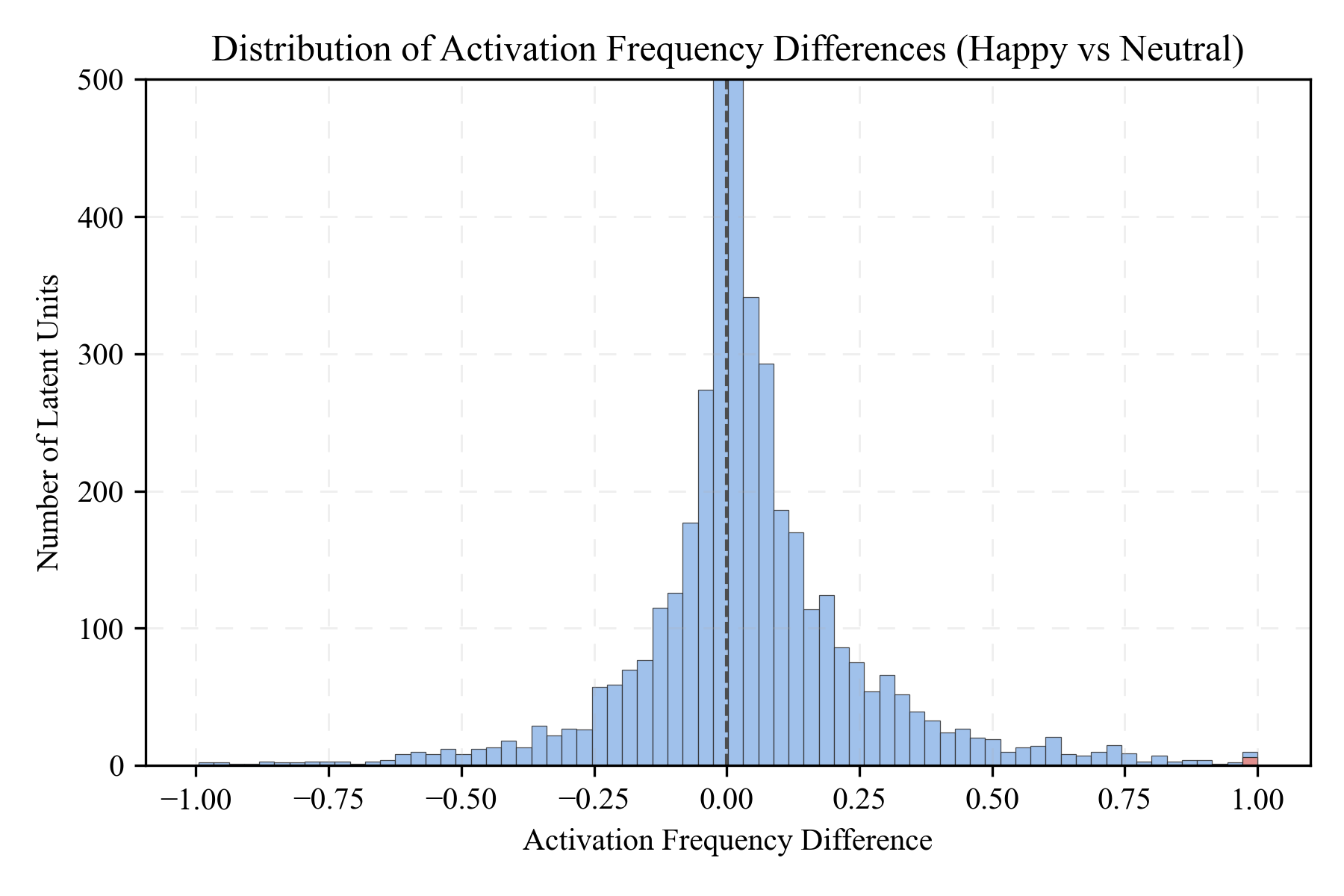}
      \caption{Happiness vs Neutral}
      \label{fig:sae-emotion-selectivity-happy}
    \end{subfigure}
    \hfill
    \begin{subfigure}{0.48\columnwidth}
      \centering
      \includegraphics[width=\linewidth]{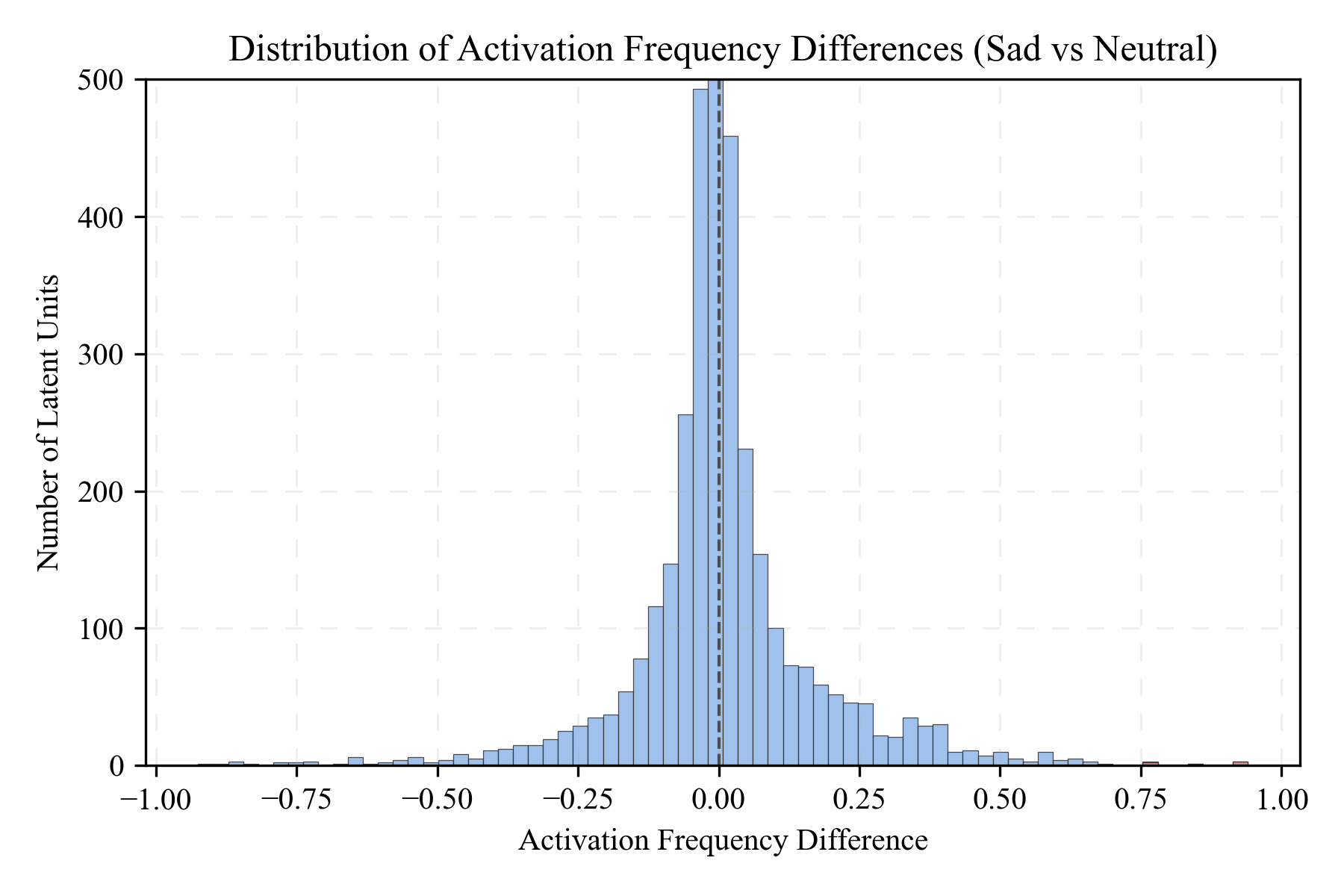}
      \caption{Sadness vs Neutral}
      \label{fig:sae-emotion-selectivity-sad}
    \end{subfigure}

    \caption{
  Distribution of selectivity score 
($\Delta_i^{(e)}$) between each target emotion and Neutral. 
Each latent feature is assigned a value reflecting how frequently it is 
activated under emotion $e$ relative to Neutral on matched texts. 
The distributions are sharply centered around zero, while only a small 
subset of latent features exhibits large positive deviations, demonstrating that 
emotion selectivity is concentrated in a sparse set of interpretable features.
  }
    \label{fig:sae-emotion-selectivity}
  \end{center}
\end{figure}

We first examine the distribution of selectivity scores for happiness and sadness, shown in Figure~\ref{fig:sae-emotion-selectivity}.
Consistent with the anger case, both distributions are sharply centered around zero,
indicating that most latent features exhibit comparable activation frequencies under emotion and neutral conditions.
Only a small subset of latent features shows substantial positive deviations,
suggesting that emotion selectivity is consistently concentrated in a limited number of sparse components across different emotions.

\label{app:emo-select}

\subsection{Single-Scalar Intensity Control}
\label{app:intensity-control}

To examine whether target-emotion intensity can be adjusted with a single steering coefficient, we vary the steering scale $\alpha_e$ from $-60$ to $+60$ while keeping the selected emotion-related latent feature fixed.
For each scale, we generate speech under matched text and speaker conditions and compute cosine similarity to the target emotion prototype in emotion2vec space.

\begin{figure}[htbp]
  \begin{center}
    \centerline{\includegraphics[width=0.5\textwidth]{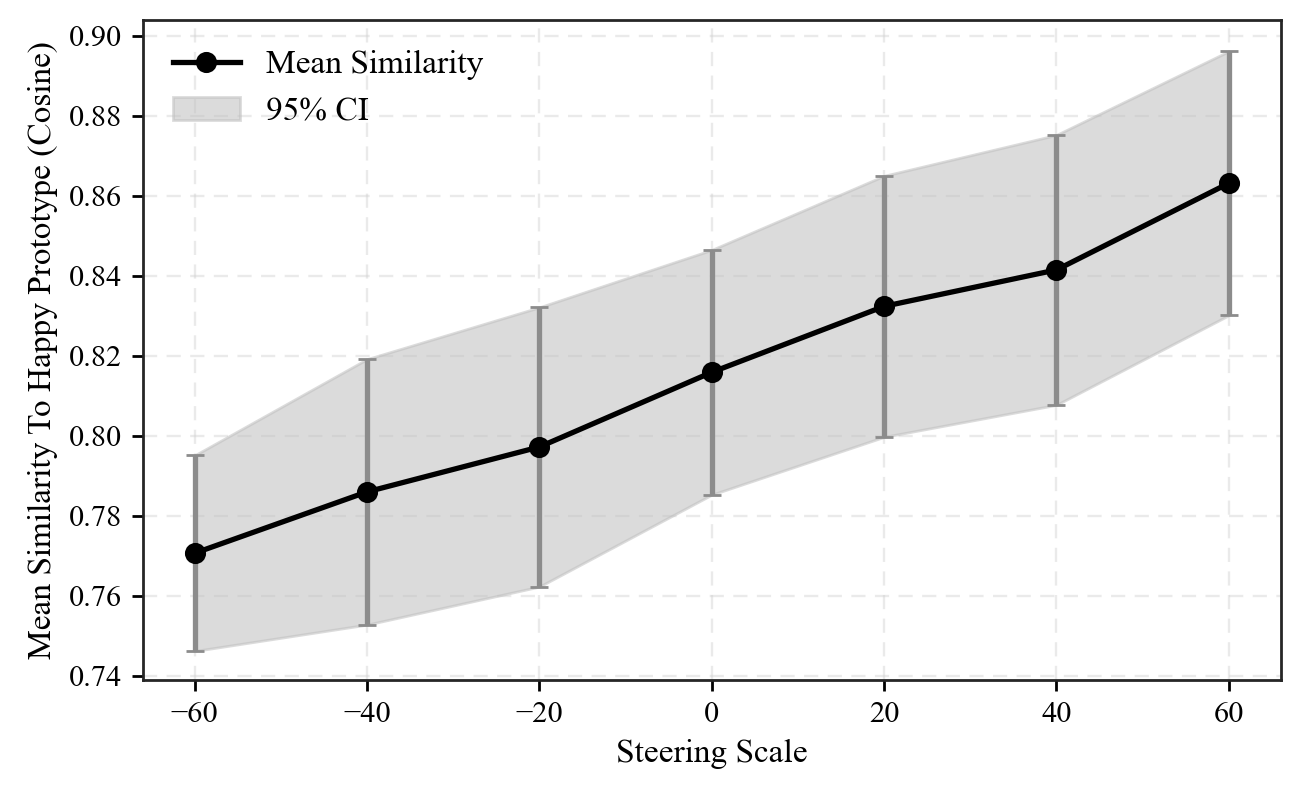}}
\caption{
Single-scalar control of target-emotion intensity.
Mean cosine similarity to the target emotion prototype is plotted as a function of steering scale $\alpha_e$.
The zero-scale condition corresponds to the neutral baseline.
Similarity increases smoothly as $\alpha_e$ increases, indicating that target-emotion strength can be adjusted through a single continuous steering coefficient.
Shaded regions denote 95\% confidence intervals.
}
\label{fig:appendix-scale-sweep}
  \end{center}
\end{figure}

As shown in Figure~\ref{fig:appendix-scale-sweep}, similarity to the target emotion prototype increases smoothly with $\alpha_e$, from approximately 0.77 at $\alpha_e=-60$ to 0.86 at $\alpha_e=+60$.
This trend indicates that the steering scale provides an intuitive control knob for adjusting target-emotion intensity.

\subsection{Cross-Emotion Behavior and Latent-Feature Budget}
\label{app:cross_emotion_latent_budget}
\begin{figure}[htbp]
\centering
\includegraphics[width=\textwidth]{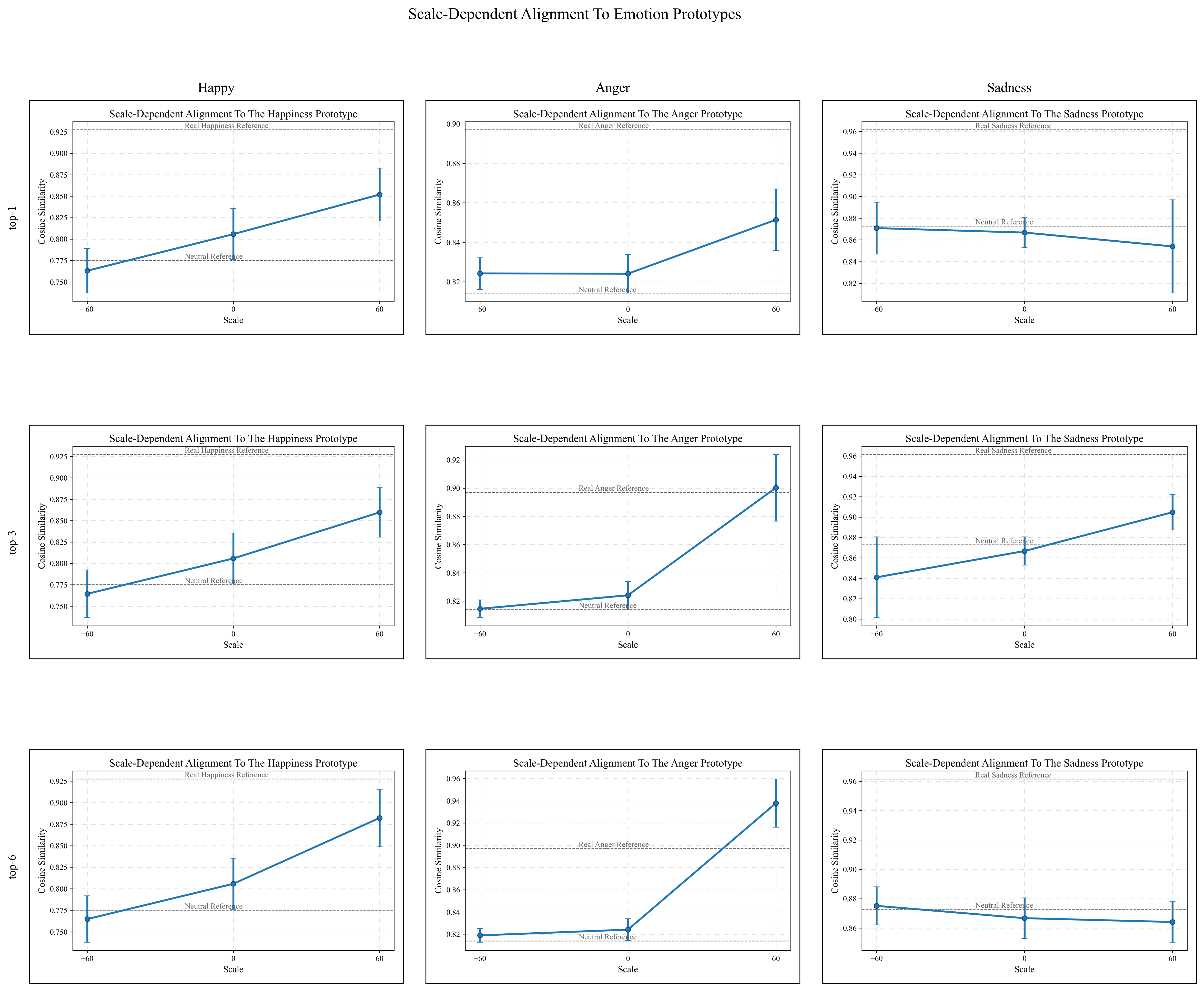}
\caption{
Scale-dependent alignment to emotion prototypes across emotion categories
and latent-feature budgets. Columns correspond to target emotions, and rows correspond
to top-$1$, top-$3$, and top-$6$ selected latent-feature settings. Similarity is calibrated
using neutral and real-emotion reference groups.
}
\label{fig:cross_emotion}
\end{figure}

We further analyze how controllability varies across both emotion categories
and latent-feature budgets. Figure~\ref{fig:cross_emotion} shows scale-dependent
alignment to emotion prototypes for happiness, anger, and sadness under
top-$1$, top-$3$, and top-$6$ latent-feature settings. Happiness shows a consistent
monotonic trend even with a single latent feature, suggesting a relatively concentrated
control direction. In contrast, anger and sadness benefit more from increasing
the latent-feature budget, indicating that their control signals are more distributed
across multiple sparse features. These results suggest that emotion-related
control is sparse but not uniformly single-feature: some emotions can be steered
by a small dominant latent feature, whereas others require a structured combination of
multiple latent features.

\subsection{Latent Feature Overlap Across Emotions}
\label{app:latent_overlap}

We examine whether the selected emotion-related latent features overlap across different emotions. Under our selection criterion (top-$m$ latent features based on the sentence-level selectivity score), we observe no overlap among the top-6 latent features across emotions in our setting.

This indicates that, at high activation levels, different emotions are largely captured by distinct sparse directions, rather than relying on a single shared latent feature. This separation supports our interpretation that emotion control is associated with emotion-specific sparse activations.

We note that partial overlap may emerge under different sparsity thresholds or selection strategies, which we leave for future work.

\subsection{Comparison with Alternative Selection Criteria}
\label{app:alternative_selection_criteria}

To examine whether the proposed sentence-level selectivity score
identifies emotion-specific latent features rather than merely frequently
activated ones, we compare it with two alternative criteria:
magnitude-based selection and token-level selection. All criteria are
computed under the same paired emotion-versus-neutral setting with
matched text and speaker conditions.

As shown in Table~\ref{tab:alternative_selection_criteria},
sentence-level selectivity yields stronger emotion alignment across
all three target emotions. This suggests that the selected latent features are
not only active, but are more reliably associated with emotion-specific
modulation under controlled conditions.

\begin{table}[htbp]
\centering
\caption{Comparison of latent-feature selection criteria for emotion induction.
Sentence-level selectivity achieves stronger emotion alignment than
magnitude-based and token-level alternatives across all three target
emotions.}
\label{tab:alternative_selection_criteria}
\begin{tabular}{lccc}
\toprule
Method & Anger $\uparrow$ & Happiness $\uparrow$ & Sadness $\uparrow$ \\
\midrule
Sentence-level selectivity & 0.912 & 0.885 & 0.880 \\
Magnitude-based selection & 0.822 & 0.820 & 0.866 \\
Token-level selection & 0.825 & 0.811 & 0.864 \\
\bottomrule
\end{tabular}
\end{table}

\section{Quantitative Acoustic Analysis}
\label{app:quantitative_acoustic_analysis}

To further examine whether the selected SAE latent feature corresponds to interpretable
acoustic changes, we measure three utterance-level acoustic features under
matched text and speaker conditions: mean F0, effective duration, and RMS
energy. For each utterance, we compare the neutral baseline with the steered
generation using paired statistics. As shown in
Table~\ref{tab:quantitative_acoustic_analysis}, steering significantly increases
mean F0 and RMS energy, while duration does not show a statistically
significant change. This suggests that the selected latent feature primarily modulates
pitch and vocal intensity rather than simply changing speaking duration.

\begin{table}[htbp]
\centering
\caption{Quantitative analysis of acoustic features under controlled steering
with matched text and speaker identity. Steering significantly increases pitch
(F0) and RMS energy, while duration shows no statistically significant change.}
\label{tab:quantitative_acoustic_analysis}
\begin{tabular}{lcccccc}
\toprule
Feature & Baseline Mean & Baseline Std & Steered Mean & Steered Std & $\Delta$ & $p$-value \\
\midrule
F0 (Hz) & 167.99 & 11.20 & 191.10 & 14.06 & +23.11 & $1.07{\times}10^{-4}$ \\
Duration (s) & 1.685 & 0.234 & 1.662 & 0.259 & -0.023 & 0.687 \\
RMS Energy & 0.02712 & 0.00354 & 0.03146 & 0.00479 & +0.00435 & 0.00769 \\
\bottomrule
\end{tabular}
\end{table}

\begin{figure}[htbp]
  \centering
  \includegraphics[width=0.5\textwidth]{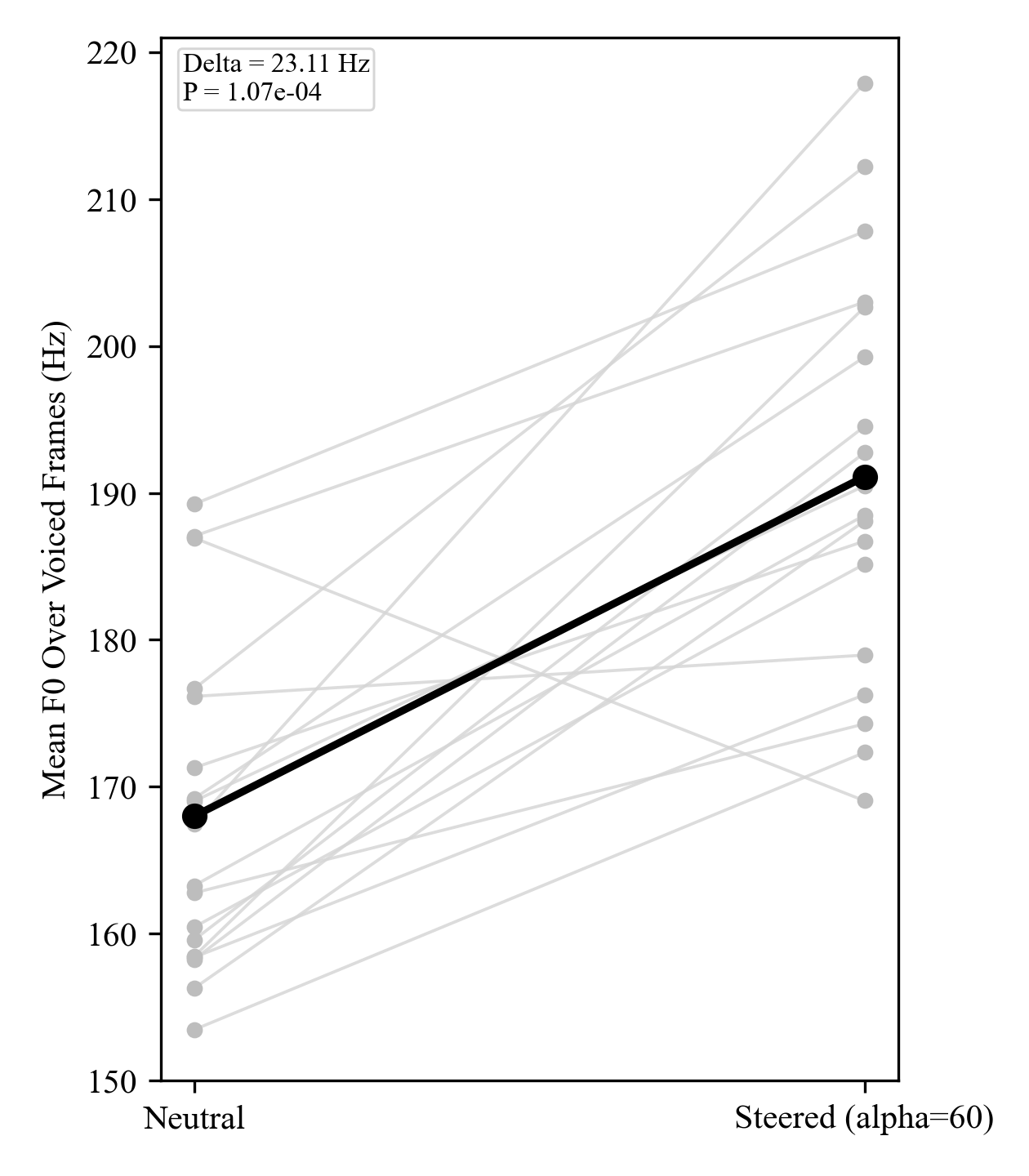}
  \caption{
Paired comparison of mean fundamental frequency (F0) between neutral and steered generations under matched text and speaker identity.
Each gray line represents one matched utterance pair, and black markers indicate condition means.
Steering increases mean F0 by $+23.11$ Hz on average ($p=1.07\times10^{-4}$).
}
  \label{fig:appendix_f0}
\end{figure}

\section{Robustness under Strong Steering}
\label{app:strong_steering}

To evaluate whether sparse steering preserves linguistic integrity under stronger
interventions, we compare SAE steering with dense global steering at a high steering
scale. As shown in Table~\ref{tab:strong_steering}, SAE steering substantially reduces
WER and deletion errors compared with global steering, suggesting that sparse
feature-level interventions introduce less decoding interference.

\begin{table}[htbp]
\centering
\caption{Robustness under strong steering. SAE steering preserves linguistic
fidelity better than dense global steering under high steering strength.}
\label{tab:strong_steering}
\begin{tabular}{lccc}
\toprule
Method & Mean WER $\downarrow$ & Mean Deletion $\downarrow$ & Max Deletion $\downarrow$ \\
\midrule
Global Steering & 2.86\% & 2.05\% & 14.89\% \\
SAE Steering & 0.57\% & 0.00\% & 0.00\% \\
\bottomrule
\end{tabular}
\end{table}

\subsection{Bidirectional Latent-Feature Intervention}
\label{app:bidirectional_intervention}

We additionally compare positive-only steering with a bidirectional latent-feature
intervention that simultaneously increases target-emotion latent features and decreases
opposing or neutral-associated latent features. This analysis is intended to
test whether sparse latent-feature interventions support more structured control than
uniform positive scaling alone.

\begin{figure}[htbp]
\centering
\includegraphics[width=0.58\textwidth]{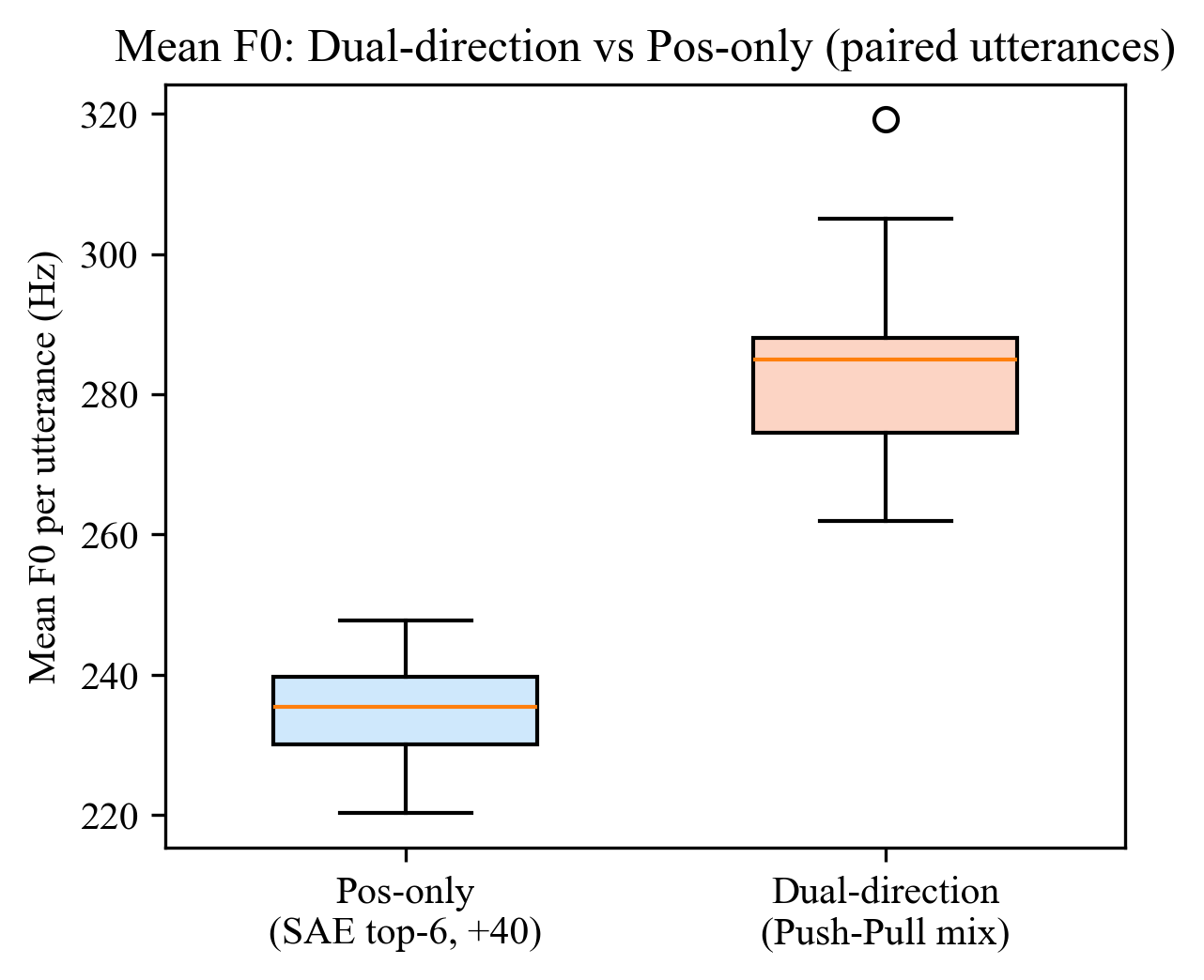}
\caption{Distribution of mean fundamental frequency (F0) comparing
positive-only steering and bidirectional latent-feature intervention under paired
utterances. Higher values indicate elevated pitch relative to the comparison
condition.}
\label{fig:bidirectional_f0}
\end{figure}

As shown in Figure~\ref{fig:bidirectional_f0}, bidirectional intervention produces
a larger increase in mean F0 than positive-only steering under paired utterances.
This suggests that sparse latent-feature control can support both reinforcement and
suppression of selected components. We treat this as a supplementary analysis
rather than the primary steering protocol used in the main experiments.

\section{Baselines and Backbone}
\label{baseline}
In this section, we present four baseline methods for comparison, as well as our backbone model, IndexTTS2, on which the SAE is trained for latent analysis and steering.
\begin{itemize}
\item VALL-E-X \cite{zhang2023speak}\footnote{\url{https://github.com/Plachtaa/VALL-E-X}} extends the VALL-E \cite{wang2023neural} framework to cross-lingual speech synthesis by modeling speech as a sequence of discrete neural codec tokens. Instead of directly generating waveforms, the model first encodes speech into quantized acoustic tokens using a neural audio codec. These tokens are then treated as a language sequence and modeled autoregressively with a Transformer-based language model.

\item Spark-TTS \cite{wang2025spark}\footnote{\url{https://github.com/SparkAudio/Spark-TTS}} proposes a single-stream decoupled speech token representation powered by BiCodec, enabling efficient LLM-based zero-shot TTS with fine-grained attribute control. Unlike multi-codebook architectures such as VALL-E-X, Spark-TTS models speech generation within a unified token sequence, improving efficiency and controllability.

\item EmoVoice \cite{yang2025emovoice}\footnote{\url{https://github.com/yanghaha0908/EmoVoice}} enables freestyle emotional prompting through natural language instructions, but emotional control remains implicit and entangled within the language model representations. In contrast, our method introduces sparse latent-feature interventions that provide interpretable and fine-grained emotional modulation at the representation level.

\item CosyVoice \cite{du2024cosyvoice1}\footnote{\url{https://github.com/FunAudioLLM/CosyVoice}} introduces supervised semantic tokens to enhance linguistic alignment in LLM-based TTS. While this improves content consistency, emotional representations remain implicitly entangled within the token embeddings. Our work further investigates fine-grained emotional disentanglement within such semantic backbones.

\end{itemize}

We adopt IndexTTS2 \cite{zhou2025indextts2} as the backbone model for our analysis due to its strong zero-shot emotional expressiveness and autoregressive semantic-token architecture. IndexTTS2 is a GPT-based text-to-semantic model that generates discrete semantic tokens in an autoregressive manner, supporting both duration-controlled and natural generation modes. It explicitly introduces emotion--speaker disentanglement, enabling independent control over timbre and emotional style via separate prompts. In addition, its soft instruction mechanism allows natural-language-based emotional guidance, making it a strong autoregressive emotional TTS system.

Emotional representations in IndexTTS2 remain implicitly encoded within the semantic latent space, without explicit structural interpretation or fine-grained modulation. This makes it particularly suitable for representation-level analysis and intervention. By building upon a well-validated emotional TTS backbone, we ensure that any observed improvements stem from our proposed sparse latent-feature steering method rather than architectural modifications, allowing us to focus on mechanistic understanding and interpretable emotional control within the semantic backbone.

\section{Baseline Evaluation Details}
\label{baseline-test-detail}

For baseline models that require natural-language instructions as emotional control signals, we provide explicit emotion descriptions following the model-specific prompting format. Each instruction ends with the special token \texttt{\textless|endofprompt|\textgreater}, which marks the termination of the control prompt in the backbone model.

The instructions used in our experiments are listed below:

\paragraph{Anger}
A person speaking in a clearly angry tone, with strong emotion and intensity in their voice, expressing anger and irritation. \texttt{\textless|endofprompt|\textgreater}

\paragraph{Happiness}
A person speaking in a clearly happy tone, with strong positive emotion and enthusiasm in their voice, expressing happiness and joy. \texttt{\textless|endofprompt|\textgreater}

\paragraph{Sadness}
A person speaking in a clearly sad and melancholic tone, with deep emotion and sorrow in their voice, expressing sadness and grief. \texttt{\textless|endofprompt|\textgreater}

\paragraph{Neutral}
A person speaking in a neutral and calm tone, with a balanced and natural voice, expressing no particular emotion, speaking in a clear and professional manner. \texttt{\textless|endofprompt|\textgreater}

In our evaluation, we directly run the publicly released open-source implementations of VALL-E-X, Spark-TTS, EmoVoice, and CosyVoice using their available checkpoints and default inference settings. Since these baseline models do not provide explicit representation-level steering mechanisms, we follow their standard control interfaces for emotional generation.

These baseline systems are primarily designed for emotional expression rather than explicit emotion suppression. As a result, they do not provide a dedicated mechanism for reducing or removing emotional attributes once introduced. Therefore, for fair comparison, we adopt a neutral-to-target emotion protocol. Specifically, for each input text and speaker reference, we first generate a neutral utterance using the neutral instruction. We then generate a corresponding utterance conditioned on the target emotion instruction (e.g., anger, happiness, sadness). Emotional consistency is evaluated by comparing the target-emotion generation against the neutral baseline, ensuring that the observed emotional shift arises from the control signal rather than speaker variation.

For speaker similarity evaluation, we compare the speaker embedding of the neutral generation with that of the emotion-controlled generation under the same speaker prompt. This protocol isolates the effect of emotional modulation while measuring whether speaker identity is preserved after emotion control. All models are evaluated using identical text inputs, speaker prompts, and evaluation metrics to ensure a fair and controlled comparison.

\section{Additional Cross-Backbone Analysis}
\label{app:llasa_analysis}

To examine whether the observed emotion-alignment behavior is specific to a single speech-generation backbone, we further evaluate LLaSA~\cite{ye2025llasa} as an additional backbone. LLaSA is a LLaMA-based speech-synthesis framework that adopts a single-layer vector-quantized codec and a single Transformer architecture, making it closely aligned with standard LLM-style autoregressive generation. This architecture provides a useful complementary setting for testing whether scale-dependent emotional alignment also appears beyond our main model.

\begin{figure}[htbp]
    \centering
    \includegraphics[width=0.6\textwidth]{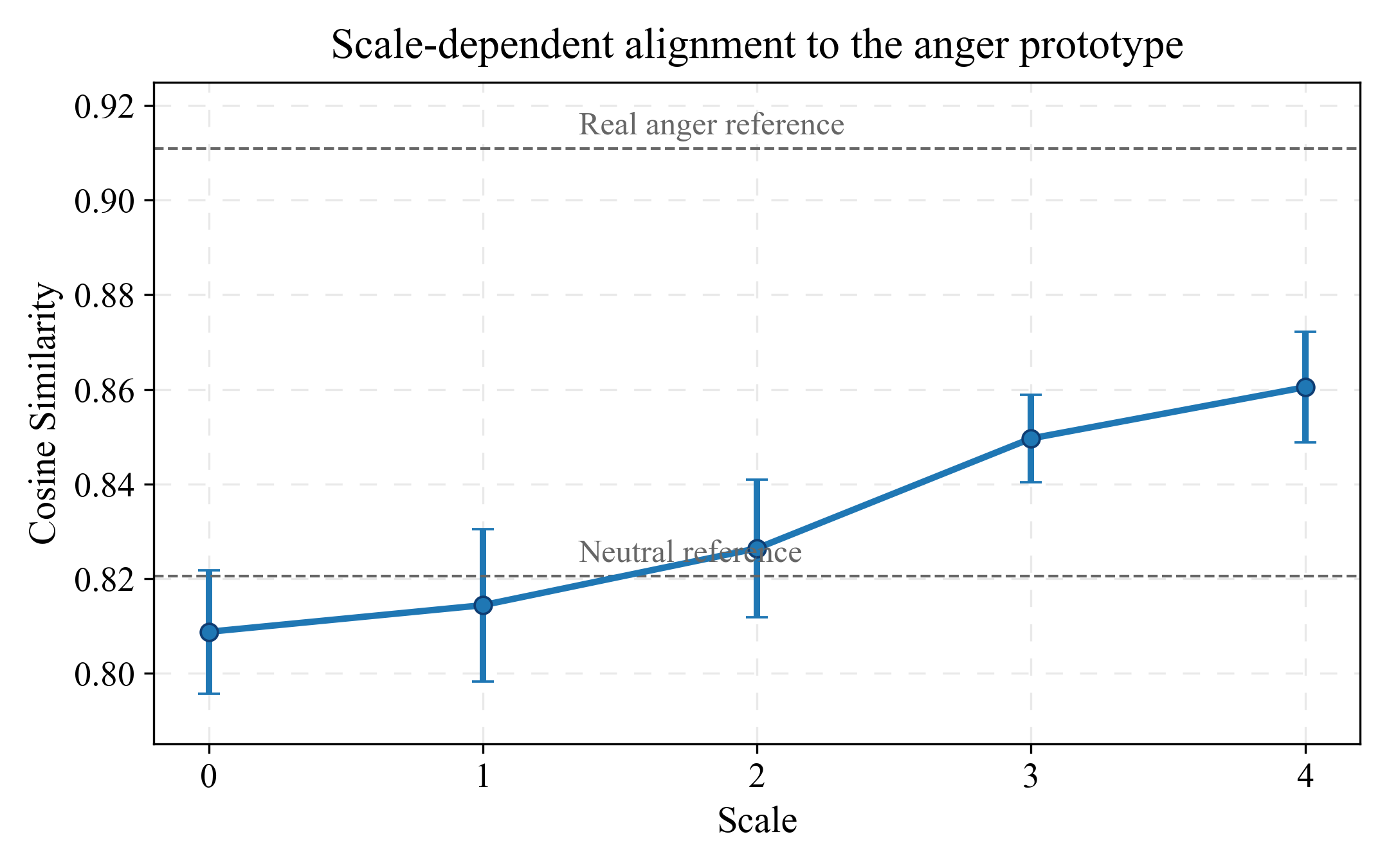}
    \caption{
Emotion-prototype alignment under top-6 latent-feature steering in LLaSA.
Cosine similarity is measured against the anger prototype; dashed lines denote neutral and real anger reference levels.
}
    \label{fig:llasa_top6_raw_cosine_alignment_0_to_4_times}
\end{figure}

As shown in Figure~\ref{fig:llasa_top6_raw_cosine_alignment_0_to_4_times}, increasing the steering scale consistently shifts LLaSA generations from the neutral region toward the anger prototype. At low scale values, the generated samples remain close to or slightly below the neutral reference. As the scale increases, the cosine similarity steadily rises, surpassing the neutral reference around scale 2 and continuing to approach the real anger reference at larger scales. This scale-dependent and monotonic behavior provides direct evidence that the intervention controls the target-emotion strength in LLaSA generations.


\end{document}